\documentclass[conference]{IEEEtran}
\IEEEoverridecommandlockouts

\usepackage{amsmath,amssymb,amsfonts}
\usepackage{comment}
\usepackage{cite}
\usepackage{hyperref}
\usepackage{graphicx}
\usepackage{textcomp}
\usepackage{xcolor}
\usepackage{siunitx}
\usepackage{tabu}
\usepackage{booktabs}
\usepackage{gensymb}
\usepackage{booktabs}
\usepackage{multicol}
\catcode`\|=12 
\usepackage{tikz}
\usepackage{pgfplots}

\newcommand{\revision}[1]{#1}

\newcommand{\fnm}[1]{#1}
\newcommand{\sur}[1]{#1}
\newcommand{\orgname}[1]{\textit{#1}}
\newcommand{\orgaddress}[1]{\textit{#1}}
\newcommand{\country}[1]{#1}
\newcommand{\email}[1]{{\small \texttt{#1}}}

\begin{document}

\title{Analysis and Perspectives on the ANA Avatar XPRIZE Competition}

\author{
\IEEEauthorblockN{Kris Hauser$^1$, Eleanor `Nell' Watson$^2$, Joonbum Bae$^3$, Josh Bankston$^4$, Sven Behnke$^5$, Bill Borgia$^6$,}
\IEEEauthorblockN{Manuel G. Catalano$^7$, Stefano Dafarra$^7$, Jan B.F. van Erp$^8$, Thomas Ferris$^{9}$, Jeremy Fishel$^{10}$, Guy Hoffman$^{11}$, }
\IEEEauthorblockN{Serena Ivaldi$^{12}$, \fnm{Fumio} \sur{Kanehiro}$^{13}$, \fnm{Abderrahmane} \sur{Kheddar}$^{14}$, \fnm{Ga\"{e}lle} \sur{Lannuzel}$^{15}$, \fnm{Jacquelyn Ford} \sur{Morie}$^{16}$,}
\IEEEauthorblockN{\fnm{Patrick} \sur{Naughton}$^1$, \fnm{Steve} \sur{NGuyen}$^{15}$, \fnm{Paul} \sur{Oh}$^{17}$, \fnm{Taskin} \sur{Padir}$^{18}$, \fnm{Jim} \sur{Pippine}$^{19}$, \fnm{Jaeheung} \sur{Park}$^{20}$, \fnm{Jean} \sur{Vaz}$^{17}$},
\IEEEauthorblockN{\fnm{Daniele} \sur{Pucci}$^7$, \fnm{Peter} \sur{Whitney}$^{18}$, \fnm{Peggy} \sur{Wu}$^{21}$, and \fnm{David} \sur{Locke}$^{22}$}\\

\IEEEauthorblockA{$^1$ \orgname{University of Illinois at Urbana-Champaign}, \orgaddress{\country{USA}},
\email{\{kkhauser,pn10\}@illinois.edu}}
\IEEEauthorblockA{$^2$ \orgname{University of Gloucestershire}, \orgaddress{\country{UK}},
\email{eleanorwatson@connect.glos.ac.uk}}
\IEEEauthorblockA{$^3$ \orgname{UNIST}, \orgaddress{\country{South Korea}},
\email{jbbae@unist.ac.kr}.}
\IEEEauthorblockA{$^4$\orgname{MACE Virtual Labs}, \orgaddress{\country{USA}},
\email{josh@macevl.com}}
\IEEEauthorblockA{$^5$\orgname{University of Bonn}, \orgaddress{\country{Germany}},
\email{behnke@cs.uni-bonn.de}}
\IEEEauthorblockA{$^6$\orgname{B98 Ventures LLC}, \orgaddress{\country{USA}},
\email{bill@borgia.us}}
\IEEEauthorblockA{$^7$\orgname{Istituto Italiano di Tecnologia}, \orgaddress{\country{Italy}},
\email{\{manuel.catalano,stefano.dafarra,daniele.pucci\}@iit.it}}
\IEEEauthorblockA{$^9$\orgname{TNO} and \orgname{University of Twente}, \orgaddress{\country{Netherlands}},
\email{jan.vanerp@tno.nl}}
\IEEEauthorblockA{$^{9}$\orgname{Texas A\&M University}, \orgaddress{\country{USA}},
\email{tferris@tamu.edu}}
\IEEEauthorblockA{$^{10}$\orgname{Tangible Research} and \orgname{California State University, Chico}, \orgaddress{\country{USA}},
\email{jeremy@tangible-research.com}}
\IEEEauthorblockA{$^{11}$\orgname{Cornell University}, \orgaddress{\country{USA}},
\email{hoffman@cornell.edu}}
\IEEEauthorblockA{$^{12}$ \orgname{INRIA}, \orgname{Université de Lorraine}, and \orgname{CNRS}, \orgaddress{\country{France}}, 
\email{serena.ivaldi@inria.fr}}
\IEEEauthorblockA{$^{13}$ \orgname{CNRS-AIST Joint Robotics Laboratory, IRL3218}, \orgaddress{\country{Japan}}
\email{f-kanehiro@aist.go.jp}}
\IEEEauthorblockA{$^{14}$ \orgname{CNRS-University of Montpellier, LIRMM}, \orgaddress{\country{France}}, 
\email{kheddar@lirmm.fr}}
\IEEEauthorblockA{$^{15}$ \orgname{Pollen Robotics}, \orgaddress{\country{France}}, 
\email{\{gaelle.lannuzel,steve.nguyen\}@pollen-robotics.com}}
\IEEEauthorblockA{$^{16}$ \orgname{All These Worlds, LLC}, 
\email{jfmorie@gmail.com}}
\IEEEauthorblockA{$^{17}$ \orgname{University of Nevada, Las Vegas}, \orgaddress{\country{USA}}, 
\email{\{paul.oh@unlv.edu, jean.chagasvas@alumni.unlv.edu\}}}
\IEEEauthorblockA{$^{18}$ \orgname{Northeastern University}, \orgaddress{\country{USA}}, 
\email{\{t.padir,j.whitney\}@northeastern.edu}}
\IEEEauthorblockA{$^{19}$ \orgname{Golden Knight Technologies}, \orgaddress{\country{USA}}, 
\email{jim@goldenknighttechnologies.com}}
\IEEEauthorblockA{$^{20}$ \orgname{Seoul National University}, \orgaddress{\country{South Korea}}, 
\email{park73@snu.ac.kr}}
\IEEEauthorblockA{$^{21}$ \orgname{Raytheon Technologies Research Center}, \orgaddress{\country{USA}}, 
\email{peggy.wu@rtx.com}}
\IEEEauthorblockA{$^{22}$ \orgname{XPRIZE Foundation}, \orgaddress{\country{USA}}, 
\email{david@xprize.org}}
}

\maketitle

\begin{abstract}
The ANA Avatar XPRIZE was a four-year competition to develop a robotic "avatar" system to allow a human operator to sense, communicate, and act in a remote environment as though physically present. The competition featured a unique requirement that judges would operate the avatars after less than one hour of training on the human-machine interfaces, and avatar systems were judged on both objective and subjective scoring metrics. This paper presents a unified summary and analysis of the competition from technical, judging, and organizational perspectives. We study the use of telerobotics technologies and innovations pursued by the competing teams in their avatar systems, and correlate the use of these technologies with judges’ task performance and subjective survey ratings. It also summarizes perspectives from team leads, judges, and organizers about the competition’s execution and impact to inform the future development of telerobotics and telepresence. 
\end{abstract}




\section{Introduction}

\begin{figure*}[tbp]
\centering
\includegraphics[width=0.9\textwidth]{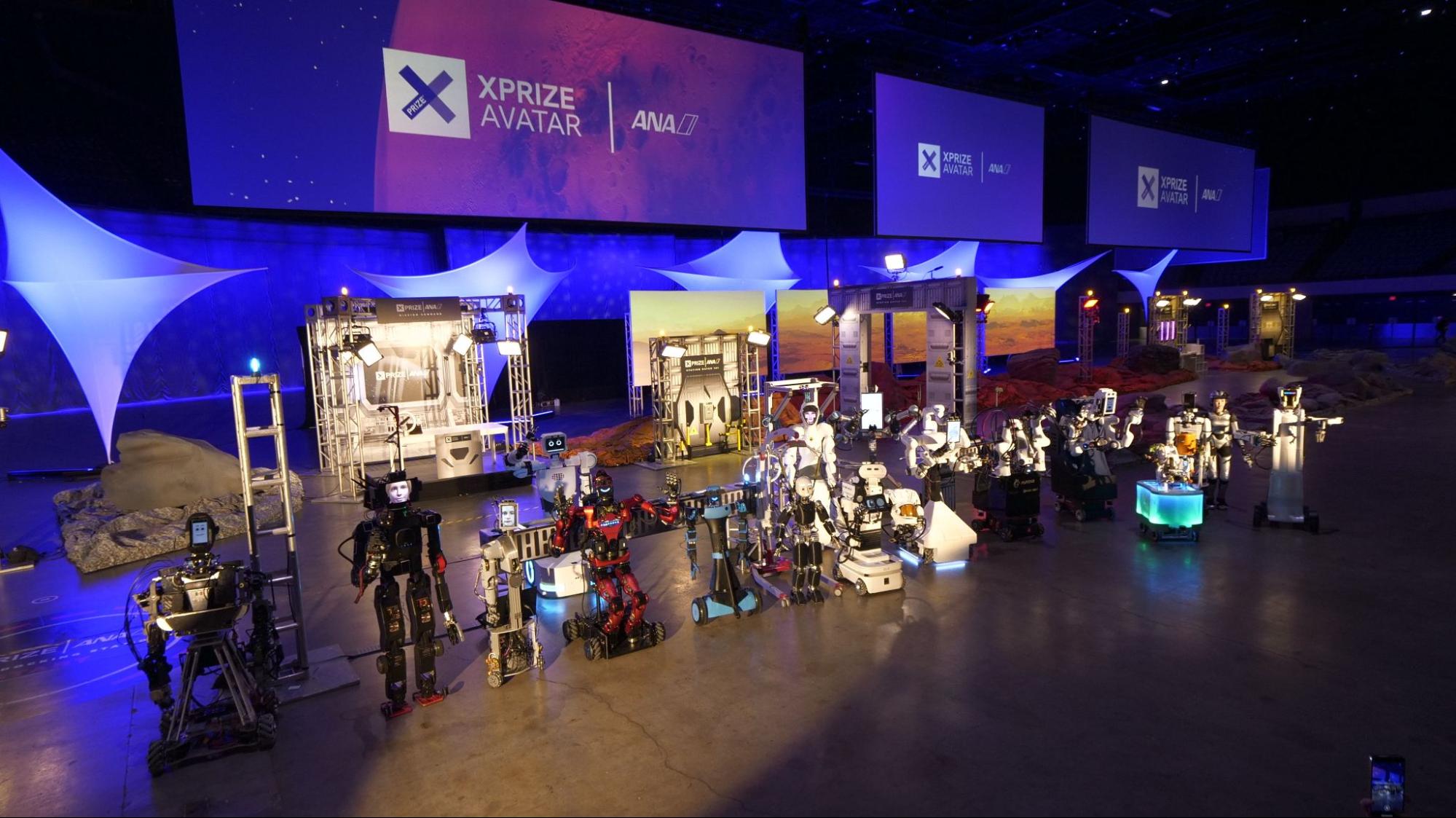}
\caption{Finalist avatar robots posed in front of the ANA Avatar XPRIZE finals course.}
\label{fig:AllFinalsRobots}
\end{figure*}

\begin{figure*}[tbp]
\centering
\includegraphics[width=0.98\textwidth]{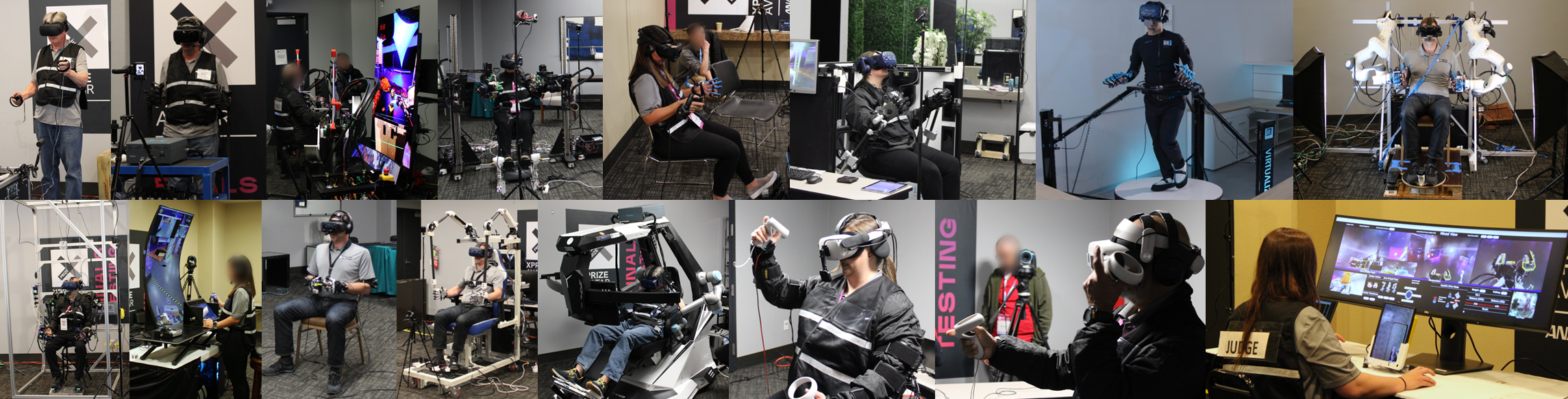}
\caption{Operator stations for the finalist teams,  representing a variety of virtual reality, 3D motion controllers, and force feedback devices.}
\label{fig:AllFinalsOperatorStations}
\end{figure*}

The ANA Avatar XPRIZE was an international competition held from 2018-2022 that challenged teams to build a telerobotic ``avatar'' system that allows human operators to transport their senses, actions, and presence across long distances. Avatars represent a next generation of telecommuting devices that enable operators to not only communicate through audio and video, but also navigate and manipulate objects in the remote environment. The potential applications of such technology are diverse and include telecommuting, emergency response, service robots, healthcare in nursing and elder care, teleoperated robots in space, and tele-tourism. Notably, this competition forced teams to consider how teleoperation could be made accessible to novice users rather than trained experts by ensuring that only judges would operate the avatar system during the competition. In September 2021, 38 teams from 16 countries competed in the XPRIZE Semifinals for a \$2 million Semifinals prize purse. 17 teams from 10 countries competed in the Finals competition in November 2022 for the remaining \$8 Million Finals prize purse (Figure~\ref{fig:AllFinalsRobots}).

A robotic avatar is a remotely controlled robotic device capable of representing a person in a location where they are not physically present.  These systems typically involve robots that are remotely-controlled by human operators via interfaces that resemble advanced virtual reality devices or vehicle cockpits (Figure~\ref{fig:AllFinalsOperatorStations}). Operators can pilot the robots to gather sensed data from and interact with remote environments. The robot can also serve as a virtual physical representation of the human in that environment in the service of telepresence applications. 

Robotic avatar systems are equipped with a variety of sensors and cameras to enable a remote operator to see, hear, and possibly feel the robot's environment. Manipulators or other tools typically also allow the operator to interact with their environment to some degree. Some robotic avatars are designed to be adaptive to a variety of tasks and environments, whilst others are specialized in specific tasks. Robotic avatars are increasingly deployed in industrial and healthcare settings to allow people to collaborate remotely. As these technologies mature, and production costs reduce, further applications are emerging, such as entertainment and business.

Telerobotics has a long history in robotics with numerous real-world applications in robotic surgery, space exploration, explosive ordnance disposal, search and rescue, and drone operation~\cite{niemeyer2016telerobotics}. Competitions like the DARPA Robotics Challenge have also allowed remote operators to control robots to complete complex search and rescue tasks involving locomotion and manipulation~\cite{Krotkov2018DRC}. In recent years, telerobotics has become more practical and accessible due to advances in networking, video transmission, and reductions in robot hardware costs. The ANA Avatar XPRIZE distinguishes itself from standard telerobotics because it emphasizes novice human users operating a robot naturally enough for it to feel like a replacement for the user’s own body. In other words, the ultimate performance objective is for operators to complete tasks as naturally and competently as being physically present in the remote space. As a result, teams in the competition explored approaches to maximize immersion and one-to-one mapping between the human’s and robot’s senses, actions, and communication capabilities. Virtual reality, full-body haptics, anthropomorphic robotic grippers, and operator assistance technologies were all employed by teams in an attempt to accomplish competition goals. Technological advancements in these areas have the potential to render telerobotics more intuitive and accessible to laypersons, provide social functions in addition to locomotion and manipulation, and enable telerobotics to be extended to new application domains. 

In this paper, we describe the organization of the ANA Avatar XPRIZE competition, the characteristics of avatar technology pursued by teams, and lessons learned from the competition. It collects varied assessments and perspectives about the competition from organizers, team leads, and judges, and analyzes the qualities of technologies that were beneficial to competition outcomes and subjective measures of presence.

The remainder of this paper is organized as follows. Section~\ref{sec:Background} provides background and related work in the field of telerobotics and telepresence, and compares the ANA Avatar XPRIZE against related robotics competitions. Section~\ref{sec:Organization} describes the organization of the competition, including rules, timelines, and teams. Section~\ref{sec:Technologies} describes the technologies used by competition teams for avatar robots, human-machine interfaces, and network communication. Section~\ref{sec:Analysis} presents the results of the competition and analyzes factors influencing competition results. In Section~\ref{sec:Discussion}, we discuss \revision{qualitative impressions and commentary from judges, and reflect on} the competition as a whole and its implications for future telepresence research and technology development.

\section{Background}
\label{sec:Background}

The term ``avatar'' is believed to have originated in Hinduism, referring to an earthly manifestation of a numinous being. The term avatar was later adopted by science fiction writers. Notably, James Cameron’s 2009 movie {\em Avatar} explores the use of cybernetically controlled biological organisms as a substrate of remote operation. Closer to the usage in the XPRIZE competition, the term ``surrogate'' was used in the 2009 movie {\em Surrogates} (based on an earlier comic series) to refer to humanoid robots that act as one's body teleoperated through immersive brain-controlled interfaces.  As a result of science fiction, the term came to refer to digital representations of people in virtual environments such as personifications in online communities and video games. The use of avatars to induce a sense of embodiment in VR settings has been shown to have various effects on users' immersiveness, physical social interaction, and spatial cognition~\cite{steed2016impact,roth2016avatar}.

More recently, the idea of using robots as avatars, or proxies, for human users to interact with the physical world remotely has been explored in various research projects. Robotic avatars are distinct from telerobotics due to a greater emphasis on cultivating a vicarious sense of presence for the operator, as well as bystanders, so that the robotic system begins to feel like an extension of the human body and its senses~\cite{aymerich-franch2016cc}. The sense of embodiment can have powerful psychological effects, and although researchers frequently pursue a direct connection between a human's sensorimotor system to similar systems on the avatar, embodiment sensations may be induced by approaches that are not so obvious. Studies in neuroplasticity have found that even in congenitally deaf individuals, areas of the temporal lobe associated with hearing are activated when subjects are viewing sign-language~\cite{merabet2005blindness}. Whether it is a biological or technically driven cause, when one modality is lacking, we compensate using input from another. For example, a study in~\cite{aymerich-franch2017jcmc} reported that users can experience haptic sensations even from a non-anthropomorphic embodied limb/agent with visual feedback alone. The haptic sensation was reported even though the setup did not include any haptic or pseudo-haptic feedback device of any form. Such findings have important implications for the understanding of the cognitive processes governing mediated embodiment, and the ANA Avatar XPRIZE has proven a strong motivator for groups to explore and showcase diverse approaches for embodiment.

\subsection{Telepresence}

Telerobotics, telepresence, and tele-existence are similar concepts that were explored starting in the 1980's~\cite{tachi1985tele}. 
Several textbooks and reviews of telerobotics and tele-manipulation demonstrate the high level of technology readiness and commercial availability of systems for both remote task execution (including for example for explosive ordnance disposal and surgery;  tEODor EVO$^\text{®}$ and Intuitive Surgical’s da Vinci Surgical System$^\text{®}$), and for remote social interaction~\cite{Avgousti2016,Lombard2015,Kristoffersson2013,Rassi2020}. 

A subset of these systems consider telepresence (the feeling of being at another location than one’s physical body) in their design, primarily focusing on vision and audition, for instance by employing a Head Mounted Display (HMD) and stereo headphones to display images from a remote platform such as a wheeled system with a built-in display screen (e.g., Suitable Technologies Beam$^\text{®}$, and Double Robotics Double 2$^\text{®}$). For instance, Opiyo~\emph{et al.} reviewed telepresence applications in dangerous environments and concluded that telepresence through sufficient visual and force feedback has positive effects on performance~\cite{Opiyo2021}. Hilty~\emph{et al.} looked at applications in clinical care and concluded that technology can significantly improve the quality of care, but also advised that ethical issues should be better explored~\cite{Hilty2020}. However, Dino~\emph{et al.} added that current robotic systems are mainly equipped with visual and auditory sensors and actuators---if present---that only have a limited capability in performing health assessments~\cite{Dino2022}.
Although several systems can indeed evoke a feeling of telepresence, transmitting only vision and audition is not very immersive or realistic and may not lead to social presence, defined as ``sense of being with another in a mediated environment''~\cite{Biocca2002} or the sense that another person is ``real'' and ``there'' when using a communication medium~\cite{Oh2018}. \revision{Tele-existence goes a step further and refers to conveying manipulation capabilities and touch sensation to the operator~\cite{tachi2015telexistence}}. The XPRIZE competition challenged the competing teams to develop systems with \revision{advanced tele-existence capabilities} to accomplish tasks and to socially interact in the remote environment, integrating advanced multisensory technologies to transmit bidirectional social cues. Van Erp~\emph{et al.} list the following important (non-verbal) social cues and how systems should implement them~\cite{vanErp2022}:
\begin{itemize}
\item Eye contact: the cameras on the telepresence robot should be close to the depicted eyes.
\item Facial expression: the telepresence robot should provide good representation and visibility of the areas around the mouth and eyes.
\item Non-verbal sounds: the system should be able to communicate non-verbal sounds in addition to speech.
\item Eye gaze and blinks: the telepresence robot must accurately display the gaze patterns and eye blinks in addition to only eye contact.
\item Gestures, movement, orientation, and posture: the telepresence robot must be able to replicate whole body signals beyond head and arm movements, e.g. waving, nodding, bowing, hand signals, inter feet distance, etc.~\cite{Anastasiou2019}.
\item Touch: the telepresence robot should be able to provide a social touch to the people in the remote environment (e.g. handshake, hug, tap on the shoulder~\cite{vanErp2015}), and to receive social touches from them.
\item Proximity, personal space: the perceived (bidirectional) proximity using a telepresence robot should be identical to that in real life.
\end{itemize}

\subsection{Robotics Competitions} 
Benchmarking robotic systems is not easy. In research labs, robots are often bespoke for a specific task environment and tested in overly favorable conditions.  To address these issues, robot competitions and challenges have proven in the last decades to be an effective way of establishing technical milestones and advancing the  robotics field~\cite{Behnke2006,nardi2016robotics}.  

Robot competitions provide task specifications, integration scenarios, and performance metrics that allow for the direct comparison of different approaches. They force participating teams to operate their systems outside their own lab, at a predefined time and under conditions controlled by the organizers. 
This induces requirements for system reliability and robustness to environmental influences. Moreover, competitions act as a catalyst for innovative ideas to spread and mature, as successful approaches are showcased amongst technical audiences, policymakers, potential investors, and to the general public. 

Notable competitions with commercial impact include the DARPA Grand~\cite{Seetharaman2006DARPAGrandChallenge} and Urban~\cite{Buehler2009darpaUrban} Challenges for self-driving cars and the Amazon Robotics Challenge~\cite{Correll2016APC} for warehouse pick-and-place.  Similar to the ANA Avatar XPRIZE was the DARPA Robotics Challenge~\cite{Krotkov2018DRC}, where human-operated robots had to solve a series of locomotion and manipulation tasks in a scenario inspired by a search and rescue mission. A data connection to a control station was available, but  most of the tasks were severely limited in bandwidth, such that teleoperation had to rely on very limited feedback. Teleoperation is also the primary mode of control in RoboCup Rescue~\cite{Tadokoro2000RoboCupRescue}, where robots explore a disaster scene and localize persons in need of help. Other RoboCup leagues, such as Humanoid~\cite{GerndtSBSB15} or RoboCup@Home~\cite{StucklerRAM2012} require autonomous robot behavior for playing soccer, and domestic service tasks, respectively. 

Some robot competitions like the DARPA Challenges and the Mohamed Bin Zayed International Robotics Challenge (MBZIRC)~\cite{BeulJFR2019, SchwarzJFR2019} support selected teams to develop their systems and also have a large prize purse. Such financial support significantly increases the interest of top research labs and leading companies to participate. While many robot competitions address mobile ground and aerial robots, some also focus on stationary manipulation tasks, e.g. the DARPA Autonomous Robotic Manipulation (ARM) program~\cite{Hackett2014DarpaARM} and the Amazon Picking~\cite{Correll2016APC} and Robotics~\cite{Schwarz2018ARC} Challenges. 

Some robot competitions like RoboCup Rescue and the DARPA Robotics Challenge also implicitly  benchmark the quality of the operator interfaces. Other competitions like RoboCup@Home include subjective scoring criteria and self-chosen tasks. One unique aspect of the ANA Avatar XPRIZE competition was that the avatar systems had to be operated not by their developers but by members of the international judging panel, who could be trained only for 45--60 minutes.  
Hence, the operator interfaces had to be intuitive.
Also, the evaluation criteria included highly subjective aspects, such as the feeling of being present in the remote space and the perception of presence of the remote operator. 

When designing the scenarios, tasks, performance metrics, and rules of a competition, the organizers must consider multiple factors. The competition must specify a set of tasks without already determining the solution. It must allow for a fair comparison of different approaches, without favoring particular technologies. It is also not easy to determine the level of difficulty. On the one hand, the tasks must be challenging and interesting to potential participants. On the other hand, it is important that they are also achievable by the best teams. Successful competitions build a community and incorporate feedback from many stakeholders to develop their rules and raise the bar with appropriate speed.

\section{Organization of Competition}
\label{sec:Organization}
XPRIZE competitions are meant to stimulate breakthroughs for ``audacious yet achievable'' technologies~\cite{Quartz2019}, and the goal of the ANA Avatar XPRIZE was to shape the future of human transportation and exploration with systems that would enable a human operator to see, hear, feel, and interact at a distance. Specifically, judges were asked to control a remote robotic avatar to complete a number of tasks to illustrate human-robot system abilities. Each challenge involved a combination of technical requirements including locomotion, communication, navigation, obstacle avoidance, fine manipulation, gripping, and conveyance of haptic sensation tasks.  Notably, the targeted user group for this challenge was {\em novice users}, in contrast to past competitions that relied on extensively trained expert pilots or members of the teams that designed and worked regularly with the robots.

This section describes the background, timeline, competition rules, participation, and engineering aspects of organizing the competition.

\subsection{Background}
\label{sec:OrganizationBackground}

The ANA Avatar XPRIZE competition was launched at South by Southwest (SXSW) in Austin, Texas, in 2018, and initially registered 99 teams from across the globe. The goal of the competition  was described in the first guidelines distributed in March 2018 as the following:

``The winning team will combine state-of-the-art technologies to demonstrate a robotic avatar that allows an untrained operator to complete a diverse series of tasks, from simple to complex, in a physical environment at least 100 km away.''

This was later adapted to focus more on the connection being made between the person operating the Avatar and the person on the receiving end. The competition organizers removed the emphasis on latency due to distance since \revision{network latency induced by 100km distance is minimal compared to other sources of system latency such as WiFi and high definition video streaming}. Instead there was a desire to focus entirely on presence. It was also considered that an entirely na\"{i}ve user would be unlikely to be proficient at using a system right away, the organizers decided to allow light amounts of training. The final competition definition was revised to the following:

``The winner of this XPRIZE will demonstrate a functional Avatar System, which consists of a human Operator controlling a robotic Avatar (Operator) at a real and/or Simulated Distance that allows the Operator to interact with another human (Recipient), or the remote environment, receiving all sensory information through the robotic Avatar. The ultimate goal is for a person (the Operator) to feel as if they are truly where the Avatar is, experiencing a sense of Presence through the Avatar.''

To implement the competition, technically-knowledgeable but na\"{i}ve users (the ANA Avatar XPRIZE Judges) were required to learn the systems and operate them to perform the tasks. As part of testing runs, each team was allotted one hour to train an operator in piloting their system. During competition runs, the team could only observe without interjecting.

\subsection{Competition Phases}

\begin{figure}[tbp]
\centering
\includegraphics[trim={5cm 2.5cm 0 5cm},clip,width=0.98\linewidth]{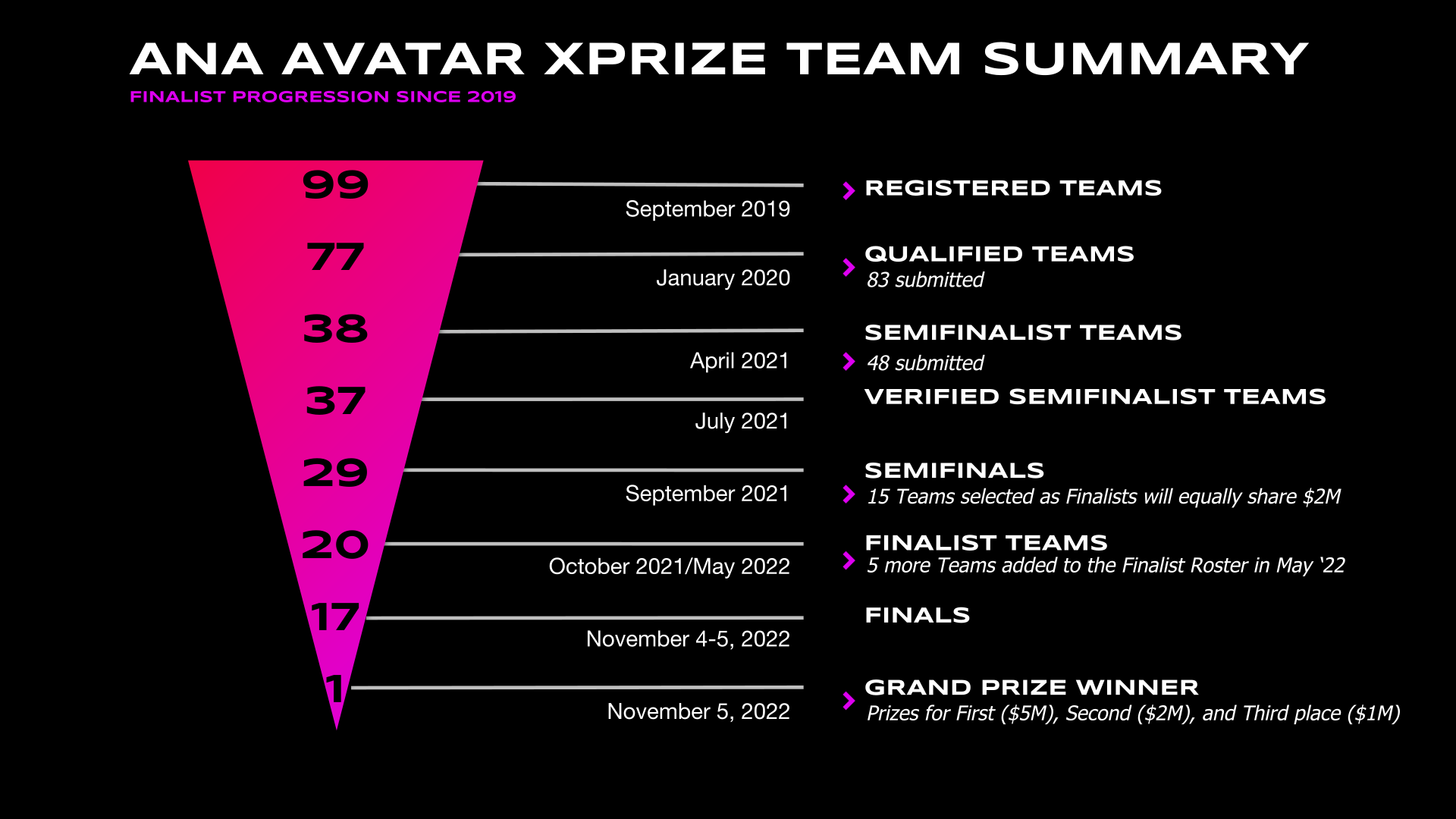}
\caption{Timeline of the phases of the competition by which teams participated and were down-selected.}
\end{figure}

The competition consisted of 5 phases: Team Registration, Team Qualification, Semifinals Qualification and Verification, Semifinals Competition, Finals Competition. This section summarizes the phases up to Semifinals Qualification and Verification. The Competition phases are detailed in subsequent sections.

\subsubsection{Team Registration}

Team recruitment was an early priority for XPRIZE and targeted plans were put into place with recruitment trips coordinated throughout the US, Europe, and Asia. Select conferences in addition to company and university visits were leveraged and later followed by team-focused webinars and workshops. The competition received interest from over 500 groups and officially registered 99 teams from 20 countries and 5 continents at the close of registration in September 2019. 

\subsubsection{Team Qualification} 
The Qualifying Submission process served as the first team technical submission, which provided the judging panel with the necessary information to evaluate the capabilities, technical maturity and potential for success of the Registered Teams. The submission consisted of questions that outlined each team’s expertise, capabilities, and plans for developing a functional Avatar System. 

In reviewing the submissions it was clear that teams were taking varied technical approaches in developing their avatar systems, as well as building for a variety of potential applications. Some competitors outfitted their systems for specific uses such as space teleoperation and medical examinations, while others developed general-purpose systems that could be applied in a range of more common situations. 

A total of 83 team submitted Qualifying Submissions were evaluated by the competition Judging Panel. After review and discussion, the Panel selected 77 Teams to advance in January 2020. Qualified teams came from 19 countries: Australia, Brazil, Canada, Colombia, Czech Republic, Finland, France, Germany, India, Italy, Japan, Jordan, Mexico, Netherlands, Russia, South Korea, Switzerland, Thailand, United Kingdom and the United States.

\subsubsection{Semifinalist Selection}
Semifinalist Selection was the second round of technical submissions required by Teams. Qualified Teams were required to demonstrate they were sufficiently advanced to progress in the Competition as a Semifinalist Team. The initial due date of the Semifinalist Selection submission was extended by 3 months to account for team impacts caused by the coronavirus pandemic and to provide XPRIZE with the necessary timetable to prepare for in-person Semifinals testing. Team submissions were evaluated by the Judging Panel who were asked to account for the technical maturity, safety, and the ability of the team to meet the competition requirements.

The Semifinalist Selection submission deadline of February 2, 2021 consisted of both written and video documentation including:
 \begin{itemize}
\item A detailed written update to the originally submitted Qualifying Submission that outlines the Team’s progress since that time.
\item A written description of a Team selected sample scenario consisting of six tasks. The tasks were required to demonstrate specific capabilities of the Avatar System that might not otherwise be showcased within the XPRIZE testing scenarios. 
\item A Video of the Avatar System demonstrating the six tasks described in the written portion of the submission from the perspectives of the Operator as well as that of the Recipient. 
\end{itemize}

In total, 48 Qualified Teams submitted proposals for reviews of which the Judges selected 38 Semifinalist Teams from 16 countries to advance.  

\subsubsection{Semifinalist Verification} 
Teams passing the Semifinalist Selection process were permitted to proceed to Semifinalist Verification in June 2021. The verification process was established to ensure teams’ physical readiness to proceed to Semifinals Testing in September 2021. It also served as an opportunity for the competition organizers to fully understand and prepare to accommodate team systems at testing. 
 
Semifinalist Verification Submissions included:
\begin{itemize}
\item Technical Enrollment: detailing weight and size, operating equipment, power requirements, network and robot diagrams, emergency shutdown procedures.
\item Personnel Enrollment: indicated the team members who would be in attendance
\item Semifinals Team Video and Documentation: demonstrating a team-devised sample scenario consisting of six tasks that best showcase system capabilities. The Scenario tasks were permitted to be the same as those devised for the Semifinalist Selection video, however it was required to be updated to show technical progress since that time. The video was later used as a portion of a Team’s Semifinals score and was devised to allow teams to demonstrate the full capability of their systems outside of the competition’s testing parameters. 
\end{itemize}

37 of the 38 Semifinalist teams submitted and passed the Semifinalist Verification process.

\subsection{Semifinals Competition}

Semifinals testing required each system to attempt three interactive Scenarios: completing a puzzle, conducting a business meeting, and exploring a museum. 
As in Finals, teams had to train Judges to operate their systems during testing, and the operator used the Avatar system to interact with another judge in the remote location. During test runs, teams were not allowed to interact with the system until the testing was completed. 

\subsubsection{Testing Procedures}

Teams were required to split their system in two locations that were separated by 300 meters and spread across different floors of the test facility. The operator system was located in a 4~m~$\times$~4~m room located on the ground floor referred to as the Operator room. Each team was provided an Operator room for the four day duration on-site and the teams set up all the equipment that was required to control their Avatar. This was also where most of the computing resources for the system were located that weren’t on the Avatar itself. The Operator room was connected via a local Avatar network to the Scenario Room. Located in the Scenario room was the Avatar itself and the Recipient Judge who interacted with the Operator Judge during the testing. The scenario rooms were approximately 3~m~$\times$~4~m and were only available to the teams for the two hour duration of their Test Slot.

Each Team was given two testing slots over the course of the two consecutive testing days. Each Slot was two hours in duration with one hour for equipment setup and Operator Judge training and one hour for the Scored Trial.  Two Judges were used for each evaluation, one judge as an operator of the avatar system, while another judge acted as the recipient.  This was a critical component of the competition as they had to interact through the avatar system in real time. In addition to task completion points, both judges scored on evaluation criteria based on their experience communicating through the system.

\subsubsection{Scoring Rubric}

Teams were scored out of 100 points. Each of the three Scenarios were worth up to 30 points while the Video submission from Semifinals Verification was worth up to 10 points. The 30 points consisted of objective and subjective criteria that the Judges evaluated after each scenario.
The 20 subjective points consisted of 12 questions of the Operator and 8 questions of the Recipient, with the rubric of Never (0 points); Rarely (.25 points); Sometimes (.5 points); Mostly (.75 points); and Always (1 points). The 10 objective points consisted of Avatar Ability and Overall System and each was scored Pass/Fail, with 1 point for each Pass and 0 points for each Fail.  The teams were evaluated on the scenarios twice over the two slots and the best score for the individual scenario was taken for the team total.

\subsubsection{Logistics and Outcomes}

In response to growing concern over the Coronavirus pandemic, Semifinalist Teams were evaluated in two groups, Plan~1, in-person in Miami, Florida in September 2021 and Plan~2, on-site testing at the team’s facilities in March/April of 2022. However, only teams that participated in Plan~1 testing were eligible for equally sharing a Semifinals milestone prize purse of \$2M. Plan~2 teams testing in Spring of 2022, were required to freeze their designs and not advance their systems after September of 2021 to keep the evaluations level. Of the 37 Semifinalist teams, 29 teams selected Plan~1 testing and were able to travel to participate in-person. 6 teams selected Plan~2 testing and hosted testing in their labs in Spring 2022. 2 teams withdrew from the competition prior to testing.

Semifinals Plan~1 testing took place at the James L. Knight Center in Miami, FL as a closed private event. This was the first opportunity for competition organizers and the expert panel of judges to meet the teams and physically evaluate their systems. In order to safely and effectively accommodate 29 teams from 11 countries participating at in-person testing, teams were split into two groupings of four days each. Teams used the first two days to move into their spaces and prepare their systems which was then followed by two days of testing. 

Semifinals Plan~2 testing evaluated the remaining 6 Semifinalist teams that were unable to travel to in-person testing in September 2021. Competition organizers and Judges traveled to 6 team-selected locations in 4 countries to evaluate their systems. Teams were required to set up their own network to a configuration provided by XPRIZE in advance and to provide two different test rooms that were separated from each other to ensure there was no communication between the two rooms other than through the avatar system.  As in Plan 1,  each team had one hour to train the Operator Judge on their system followed by the Scored Trial. However, instead of a day between Test Slots, the Judges would switch roles after the Scored Trial and immediately conduct the 2nd Test slot. 

In May 2022 with Semifinal Testing completed, XPRIZE announced the 20 teams advancing to Finals. 15 teams came from Plan 1 testing, with the remaining 5 teams coming from Plan 2 testing. All advancing teams earned a score of 80~pts or higher.

\subsection{Finals Competition}

\begin{figure}[tbp]
\centering
\includegraphics[width=0.5\textwidth]{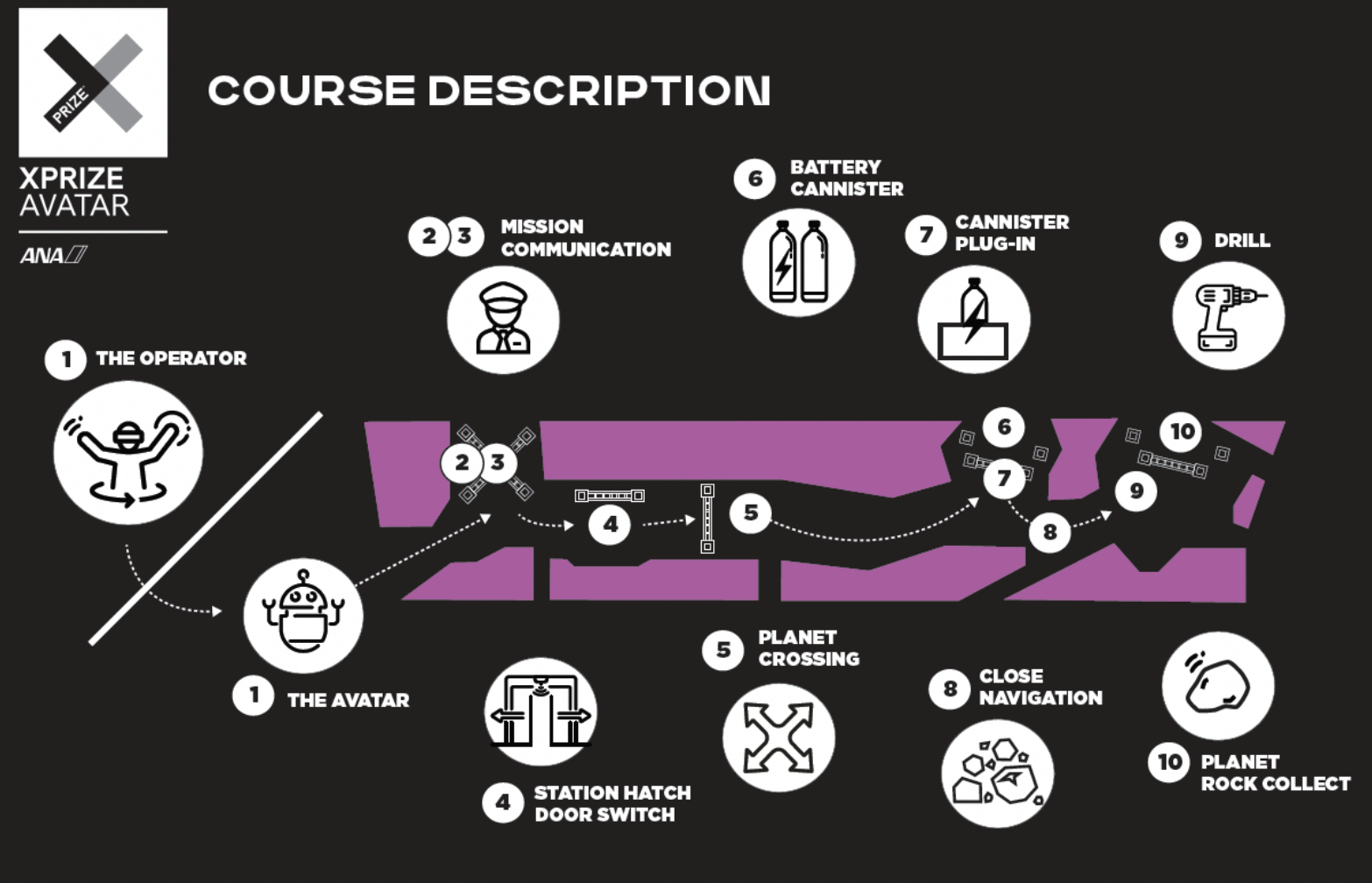}
\caption{Schematic of the finals course as provided by XPRIZE.}
\label{fig:FinalsCourse}
\end{figure}

After analyzing the results of Semifinals, it was learned that the teams moving on to Finals clearly demonstrated the ability to create a combined experience between Operator and Recipient Judges. So in order to create a test that would showcase diverse capabilities and the future potential of avatar systems, the emphasis on the remote experience decreased in the evaluation criteria. The result was a 40 meter long test course that simulated a mission on the surface of an unnamed planet focusing on the avatars ability to complete a series of challenge tasks (Fig.~\ref{fig:FinalsCourse}).

\subsubsection{Tasks and Scoring Rubric}

Ten tasks were designed to test the avatar's ability as listed in Table~\ref{tab:FinalsTasks}, with the intention that tasks increase in difficulty as the avatar progresses across the course. If the avatar was not able to complete the task, the run was terminated. A 25 minute time limit was also imposed on each run.   Each task was scored Pass/Fail, with 1 point for each Pass and 0 points for each Fail. 

\begin{table*}[tbp]
    \centering
    \renewcommand{\arraystretch}{1.2}
    \footnotesize
    \begin{tabular}{@{}p{0.6cm}>{\raggedright\arraybackslash}p{1.8cm}>{\raggedright\arraybackslash}p{5cm}>{\raggedright\arraybackslash}p{1.8cm}>{\raggedright\arraybackslash}p{5cm}@{}}
        \toprule
         {\bf Task} & {\bf Name} & {\bf Description} & {\bf Capability} & {\bf Scoring Rubric (1 point)} \\
         \midrule
         1 & Initiation &  An Operator remotely connects to the Avatar robot, and maneuvers to the mission control desk & Basic mobility & Was the Avatar able to move to the designated area? \\
        2 & Mission communication & The Avatar reports to the Mission Commander and introduces themself & Audio and video &  Did the Avatar introduce themself to the mission commander? \\
        3 & Mission communication & The avatar receives the mission details and confirms them with the Mission Commander & Audio and video & Was the Avatar able to confirm (repeat back) the mission goals? \\
        4 & Station hatch door switch & The avatar activates a switch which opens the station door & Grasping & Was the Avatar able to activate the switch? \\
        5 & Planet crossing & The avatar exits the mission control room through the door and travels across the planet to the next task & Advanced mobility over distance & Was the Avatar able to move to the next designated area? \\
        6 & Battery canister & The avatar must identify the full battery canisters that are amongst empty canister & Ability to identify weight & Was the Avatar able to identify the heavy canister? \\
        7 & Canister plug-in & The avatar places the correct canister into the designated slot which triggers the lighting of the next task zone & Manipulation & Was the Avatar able to lift up and place the heavy canister into the designated slot? \\
        8 & Close navigation & The avatar navigates along the planet's surface to arrive at the next task & Navigation and mobility & Was the Avatar able to navigate through a narrow pathway to get to the designated area? \\
        9 & Drill & The avatar must use the drill to remove the door & Advanced manipulation & Was the Avatar able to utilize a drill within the domain area? \\
        10 & Planet rock collect & The avatar must reach through the barrier to identify the rough textured rock and retrieve it & Haptics & Was the Avatar able to feel the texture of the object without seeing it, and retrieve the requested one?  \\
         \bottomrule
    \end{tabular}
    \caption{Finals task description and rubric. If an Avatar is unable to complete a task, the run is terminated.}
    \label{tab:FinalsTasks}
\end{table*}

Additionally, Operator and Recipient judges scored their experience of feeling present in the remote space and of perceiving the presence of the remote operator, respectively. The Operator Experience  category was evaluated based on the following criteria (up to 3 points): 

\begin{itemize}
    \item The Avatar System enabled the Operator Judge to feel present in the remote space and conveyed appropriate sensory information.
    \item The Avatar System enabled the Operator Judge to clearly understand (both see and hear) the Recipient.
    \item The Avatar System was easy and comfortable to use.
\end{itemize}

The Recipient Experience category was evaluated based on the following  criteria (up to 2 points):

\begin{itemize}
    \item The Avatar Robot enabled the Recipient Judge to feel as though the remote Operator was present in the space.
    \item The Avatar Robot enabled the Recipient Judge to clearly understand (both see and hear) the Operator.
\end{itemize}

Each subjective question was worth up to 1 point and was scored in 0.5 point increments according to the rubric ``Never / Poor'' (0 points), ``Sometimes / Fair'' (0.5 points), and
``Always / Good'' (1 point).

Each team was allotted a testing run on each of 2 testing days, and the best score was recorded as the team's final score. In order to qualify for the top prize, a team was required to complete the entire 10 task course. Point total ties were broken by time of the last completed task, which meant that speed was indeed an important factor in ranking the top-scoring teams.

\subsubsection{Logistics}

In June 2022, teams were provided parameters of the finals course and logistics, including tasks, scoring criteria, and sample objects to be manipulated. It should be noted that 5 months were allotted for teams to modify their entries to complete the tasks on the sample objects, which were indeed very similar to the objects used in the Finals course.

The Avatar XPRIZE Finals were held in November 2022 in Long Beach, California in the arena shown in Figure~\ref{fig:AllFinalsRobots}. 17 of the 20 Finalists teams participated in the event (which included the merger of two teams). Finals testing was open to the public and was broadcast via live video stream. This was the first publicly attended testing event conducted by XPRIZE since the Ansari XPRIZE in 2004.  Around 2500 people attended in person across the two days of the competition, and almost 10,000 unique viewers watched online.

Similar to Semifinals, teams were provided a garage space and two days to prepare their systems. These garages also doubled as demo spaces for the public who were able to observe the teams preparing in their garages. The biggest physical difference to Semifinals testing was that the Operator Rooms at Finals were only available to teams approximately 90 minutes before their test runs. In this configuration,  teams had to adapt their systems to be moved and set up quickly to allow enough time for operator training before their test run began. Because testing was conducted in front of a live audience, XPRIZE pushed teams to keep to the overall program schedule. However, due to the lack of options for Operator rooms and the unique requirements of each team’s system, doing this efficiently proved to be a logistical challenge for both the Teams and XPRIZE. Ultimately, teams managed to be flexible and creative and all teams were able to test accordingly.

The only communication allowed between the Operator Room and the Avatar robot on the course was provided via the Avatar network. The Avatar network was a connection XPRIZE provided from each team’s Operator room through the building and connected wirelessly to the Avatar robot on the course. In addition to being free from network cables while on course, teams also had to be power independent and run on batteries only during their test runs.

In order to familiarize teams with the course and the network, each team was provided a qualifying run on the course. As long as the team received a minimum score of 4 points they received a test slot for Day 1 testing.  All 17 competing teams passed the minimum qualification score and advanced. Even though Qualifying scores didn’t count for overall scores, Day 1 test runs were ordered in the rank of their Qualifying run score with the highest scoring team running last. However, one of the 17 teams' robots was damaged during their Qualifying round and wasn’t able to run during Day 1. The other 16 teams did run on the course with the top 12 teams advancing to Day 2 testing. The best score on either of the scored days was used as the team’s final score.

\subsection{Judging} 

Judges needed to be knowledgeable in a range of technical areas, as well as able to reliably act in a fair and impartial manner. Judging roles included:

\begin{figure}[tbp]
    \centering
    \includegraphics[width=0.96\linewidth]{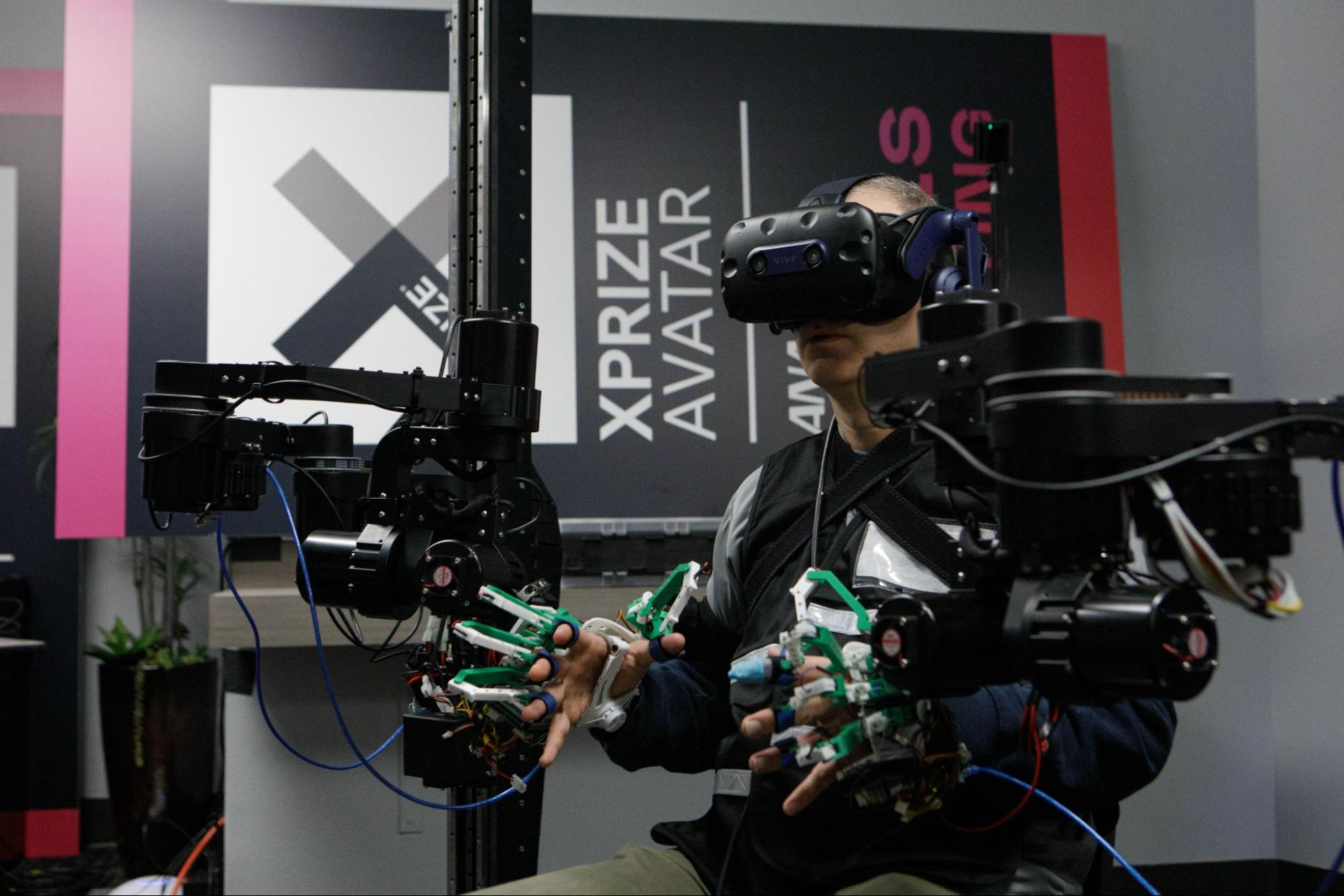}
    \caption{An Operator judge controls the Team SNU avatar robot using a head-mounted display (HMD), force feedback arms, and haptic gloves during Finals.}
    \label{fig:OperatorJudge}
\end{figure}

{\bf Operator.}
Operates the Avatar System in a remote Operator Room during test runs (Figure~\ref{fig:OperatorJudge}). Responsibility to actively listen and comprehend the explanations provided by team members and to subsequently operate the robots effectively. This necessitated a condensed one hour training period to acquire an understanding of the robot's capabilities and how to navigate it through the course. Additionally, the Operator had to adapt to the differences between physical actions and their execution through the robot's interface. revision{Due to the common use of VR equipment amongst teams, operator judges were required to identify themselves as being not prone to VR sickness.}

\begin{figure}[tbp]
    \centering
    \includegraphics[trim={0 4cm 0 0},clip,width=0.96\linewidth]{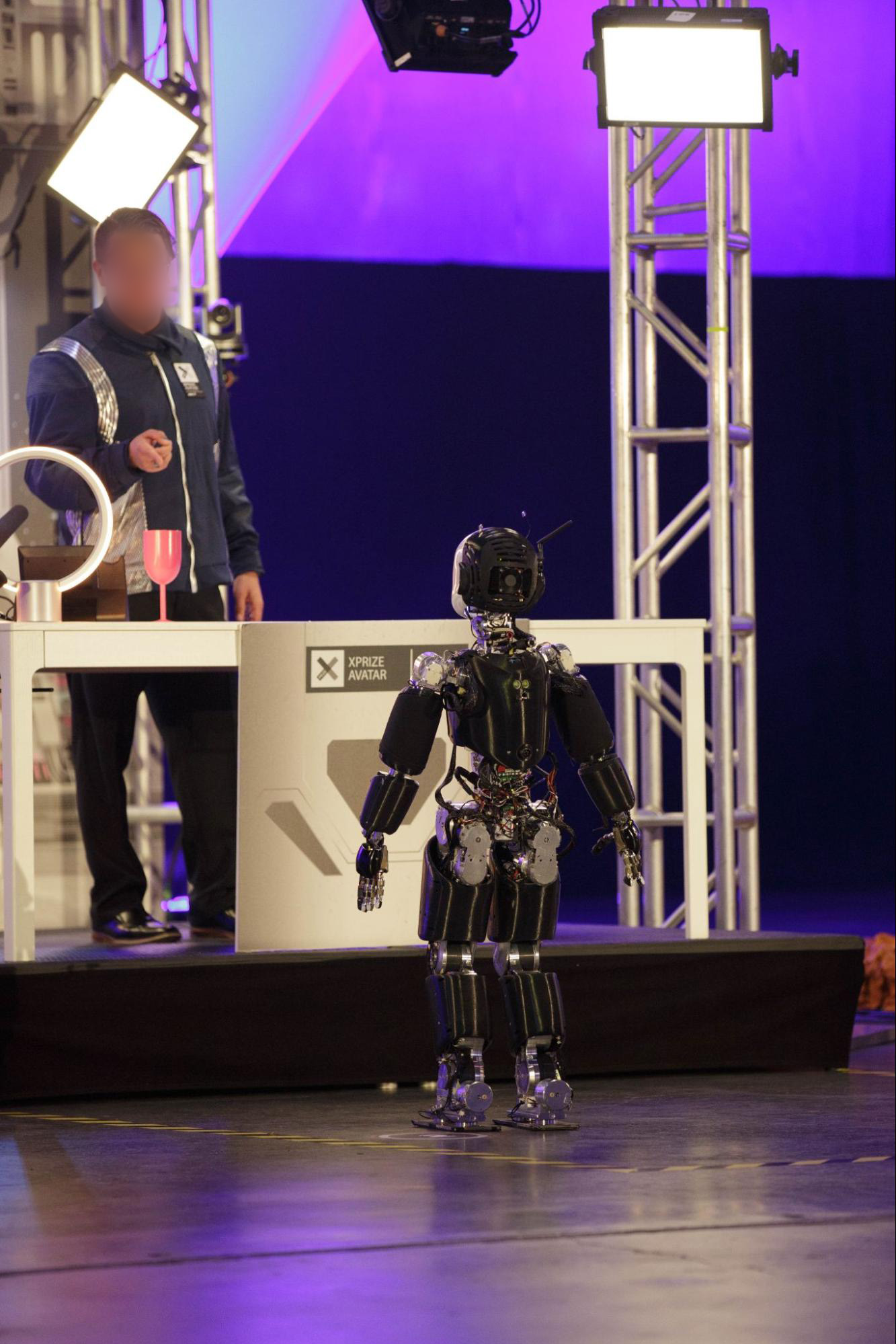}
    \caption{A Mission Commander judge corresponding with the Operator through the iCub avatar robot during Finals.}
    \label{fig:MissionCommander}
\end{figure}

{\bf Mission Commander (Recipient).}
Responsibility is to communicate to the operator judge through the avatar robot and to ensure that audio and visual communications were functioning properly (Figure~\ref{fig:MissionCommander}) . Additionally, commanders were responsible for a subjective judgment about whether the robot's visual and audio capabilities supported communication of instructions, and whether the remote human operator was perceived as ``present" in the test environment.

{\bf Scoring.}
Responsibility for monitoring and overseeing the live video feeds from the operators, the robot, and the cameras in the arena. This role included ensuring that tasks were completed in accordance with the established rules and regulations, and subsequently updating the official score and arena graphics by verifying and calling ``Point."

{\bf Staging.}
Responsibility to prioritize safety by ensuring that all necessary emergency stop measures were in place and that there were no potential hazards such as tangled or loose cables. Additionally tasked with guiding the team and their avatar to the designated start position in a fair and orderly manner  and to confirm that the course was clear and that the operator was prepared for the test run to begin.

{\bf Course Manager.}
Responsibilities included ensuring that all test elements of the course were properly positioned before a team could begin their run and monitoring the course and related machinery for any abnormalities or deviations that may warrant an emergency stop. These tasks were crucial to the smooth and safe operation of the course.

\revision{
{\bf Addressing the Subjectivity of Judging.}
} One of the biggest challenges for the organizers of the Avatar XPRIZE was the balance of evaluating subjective and objective criteria and enforcing fairness of subjective evaluations during testing events. Avatars are unique from other robots in that it is important to understand the sense of presence or connection between the Operator and the environment they are interacting with, including other humans within the remote environment. Although subjective evaluations of performance are routinely evaluated in human-robot interaction (HRI) studies, XPRIZE was not able to find prior examples of a robotics competition that blended evaluation of objective and subjective elements. Moreover, because it would be impractical for a single judging panel to evaluate every avatar system due to fatigue over the long duration of competition events, it was logistically necessary for each system be evaluated by a different set of Judges. Since each trial presented a unique experience for the Operator Judge, it was important to consider fairness and consistency between runs across dozens of teams. \revision{Inter-operator variability, in terms of preexisting operator skill, experience with similar systems, ergonomics, and fatigue, indeed affect the variability in performance of a judge operating an avatar system. Some variability was considered to be helpful, in that teams were forced to consider variability in operator populations to develop robust, intuitive interfaces and training regimes.  Moreover, results from the semifinals competition were used to help estimate judge proficiency, and these estimates were used to construct a balanced judge assignment for the final competition so that each team was assigned two operator judges whose average proficiency was roughly consistent.}

\subsection{Engineering and Safety}

Consistency of connectivity can become an issue for competitions involving communication over wired and wireless networks. To achieve tight control over network conditions, XPRIZE built and maintained a competition network to host all team network communications during both Semifinals and Finals testing events.  Communications between the Operator and the Avatar were only permitted through the Network to ensure a fair and controlled testing environment.   Each team's Avatar System was required to communicate over the Competition Network between two locations: the Operator Control Room and Test Room (Semifinals)/ the Test Course (Finals). These two locations were separated by a significant distance within the same building but were connected by a wired network. Internet access was provided through the Competition Network. Furthermore, the Competition Network was available for the use of teams for the duration of their time on site at Semifinals and Finals. 

Although the Semifinals allowed the Avatar robot to be connected to the network with a wired connection, robots were required to be tetherless in the Finals. Future avatars will need communications to be highly robust and well-tested in a range of environments.

In order to facilitate safe testing, wireless E-Stops were required for operation at Semifinals and Finals. Moreover, teams needed to provide control functionality to the Operator to place the Avatar robot in a safe mode via  the XPRIZE Competition Network.  During all testing, the E-stop and the safe operation of the Avatar was monitored by a designated Team member, who was stationed within line of sight of the robot at all times. Moreover, the Finals course was designed in a way to safeguard any humans from avatar robot malfunction. 




\subsection{Teams}

The teams that participated in the Avatar XPRIZE ranged in nature and composition and included university groups, commercial companies, leading research labs and unaffiliated individuals. Nearly 100 teams were involved from 5 continents and over 20 countries. Teams ranged in size from a single member to groups consisting of over 50 personnel and in age from current high school students to teams consisting of retired scientists.  Table~\ref{tab:TeamEnrollment} shows the breakdown of team distribution and sizes. It is evident that teams at semifinals were significantly larger than at earlier phases, and this could be either due to smaller teams dropping out, teams merging, or teams recruiting new members.  

\begin{table*}[tbp]
\centering
\includegraphics[width=0.8\textwidth]{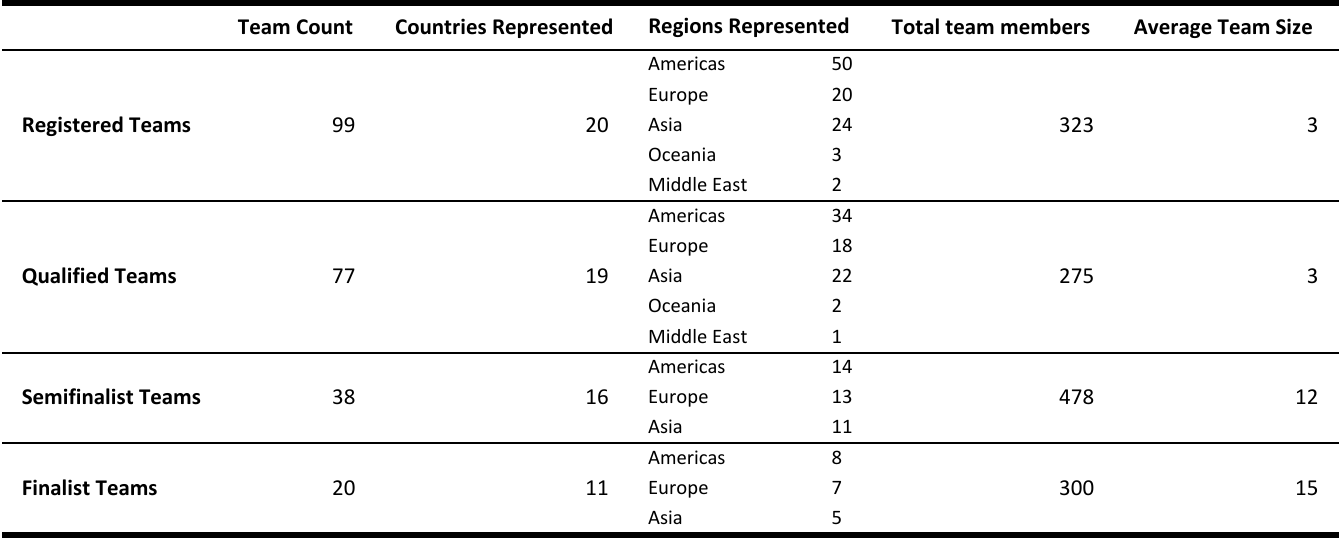}
\caption{Summary of team enrollments.}
\label{tab:TeamEnrollment}
\end{table*}

The competition faced similar gender diversity challenges as other competitions in the robotics field, with approximately 10\% of team members being female throughout all phases of the competition. Fewer than half the teams at Semifinals and Finals included female participation.

\section{Avatar Systems}
\label{sec:Technologies}
\subsection{Challenge Requirements}
Guidelines of the competition defined an avatar system as consisting of an avatar robot and operator station, connected by a communication network, that allows a human operator to sense, act, and communicate in the avatar’s environment. Moreover, the operator should have a sense of immersion in the remote environment, and learning to operate the system should not require an extended training period. Notably, the avatar robot needs sensors to replicate much of the human sensory system and actuators to replicate many of the human musculoskeletal system; the operator station needs to present a human-machine interface that renders remote sensing data to the operator and transfers the operator’s commands to the remote robot in an interpretable and intuitive fashion; and the communication network and infrastructure should support reliable and low-latency transfer of data between the two endpoints. Avatar systems should also facilitate social capabilities for communication, interaction, and social connection between operators and ``recipients'' – humans interacting with the operator but with the robot as intermediary.   We note that these social capabilities were stated in the original goals for the competition but were de-emphasized by XPRIZE during the Finals, for reasons later discussed in Section~\ref{sec:SemifinalsToFinals}.

\subsection{Avatar Robot Technologies}

We begin by discussing the technologies that comprise avatar robots used by XPRIZE finalist teams, focusing on the navigation, manipulation, vision, sensing, and actuation aspects (Table~\ref{tab:RobotTechTable}).  Avatar robots were required to be mobile and operate in human environments, which necessitated human-like size, sensing, and actuation characteristics.  XPRIZE required robots to weigh no more than 160\,kg and stand no more than 180\,cm in height, and during finals had to operate under battery power for at least the 25 minutes designated for the test run

\begin{table*}[tbp]
    \centering
    \includegraphics[width=0.96\linewidth]{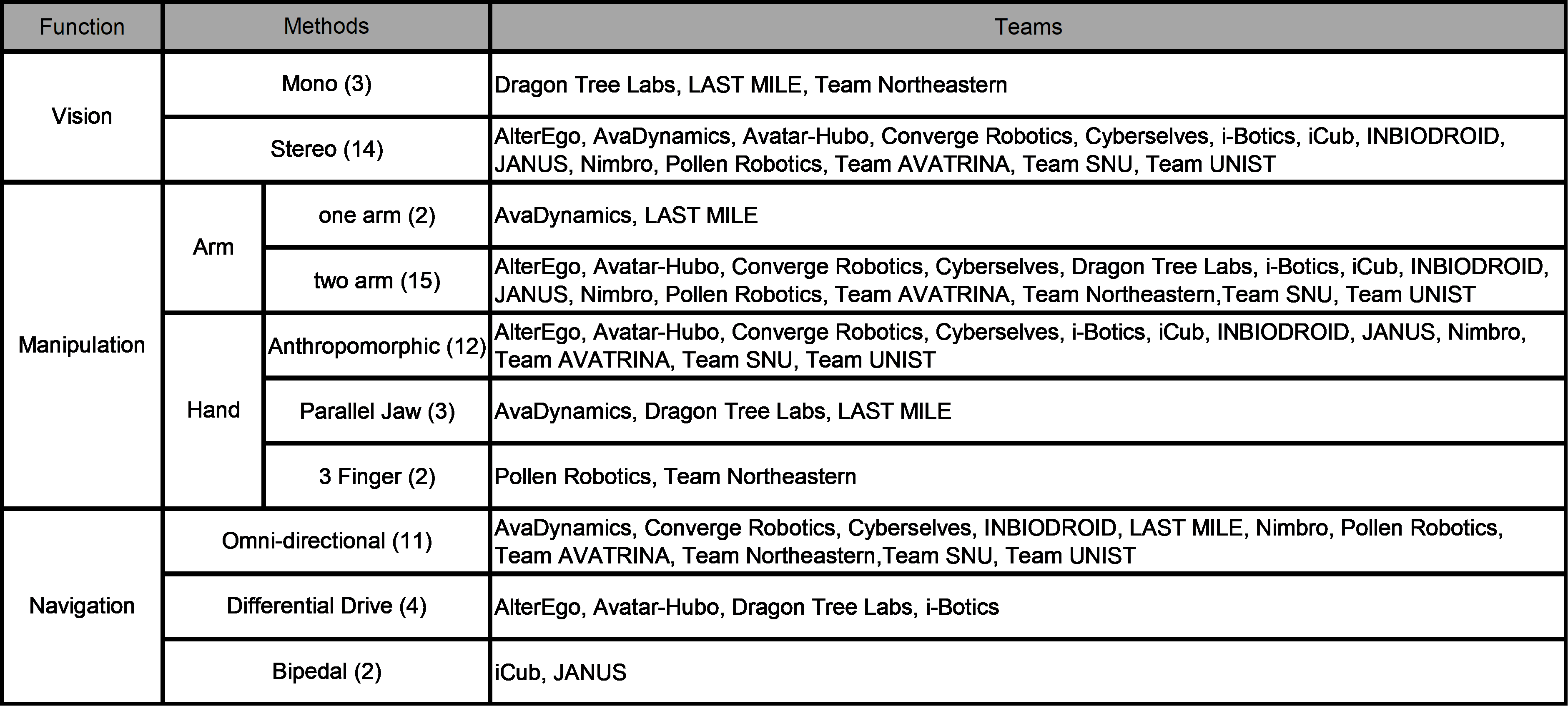}
    \caption{Breakdown of robot technologies used by teams}
    \label{tab:RobotTechTable}
\end{table*}

\subsubsection{Navigation}

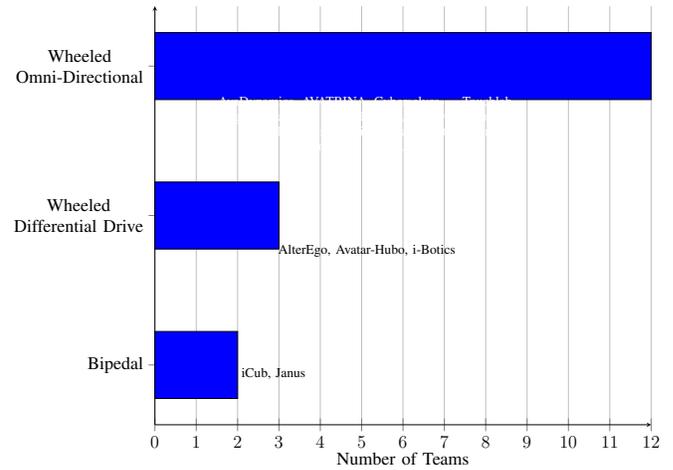
\begin{figure}
    \centering
     \resizebox{\columnwidth}{!}{
    \begin{tikzpicture}
    \begin{axis}[
        axis y line       = left,
        axis x line       = bottom,
        tickwidth         = 5pt,
        enlarge y limits  = 0.2,
        xmin              = 0,
        xmax              = 12,
        width             = 1.5\columnwidth,
        yticklabels={
            Wheeled\\Omni-Directional,
            Wheeled\\Differential Drive,
            Bipedal
        },
        ytick             = data,
        bar width         = 45pt,
        xlabel            = Number of Teams,
        xlabel style={
        font =\large,
        align=center
        },
        major grid style  = {
            line width = 0.2pt, 
            draw=gray!50
            },
        ymajorgrids=false,
        xmajorgrids=true,
        xminorgrids=false,
        major tick length=3mm,
        minor tick length=2mm,
        y tick label style={
        font =\large,
        align=center
        },
        x tick label style={
        font =\large,
        align=center
        },
    ]
    \addplot[xbar,fill=blue] coordinates {
        (12,2)
        (3,1)
        (2,0)
    };
    \end{axis}
    \node[font=\small, text=white, align=center] at (5,7.1) {AvaDynamics, AVATRINA, Cyberselves | Touchlab, \\Dragon Tree Labs, INBIODROID, Last Mile, NimbRo,\\Pollen Robotics, Tangible, Team Northeastern,\\Team SNU, Team UNIST};
    \node[font=\small, text=black, align=center] at (5,4.1) {AlterEgo, Avatar-Hubo, i-Botics};
    \node[font=\small, text=black, align=center] at (2.8,1.2) {iCub, Janus};
    
    \end{tikzpicture}
    }
    \caption{Taxonomy of the avatar locomotion types used during Finals.}
    \label{fig:LocomotionTypes}
\end{figure}

Navigation is a fundamental ability of avatar robots, and the 17 Finalist teams provided their avatars with the ability to navigate in different ways. Figure~\ref{fig:LocomotionTypes} provides a taxonomy of the adopted locomotion types. In particular, 15 teams adopted wheeled locomotion, 12 of which used omnidirectional bases with at least three Mecanum wheels, thus providing the Avatar the capability to move forward, backward, laterally, and rotate freely. The other three teams using wheeled locomotion, namely AlterEgo, Avatar-Hubo, and i-Botics, implemented differential drive locomotion, with only two actuated wheels placed with a parallel axis of rotation. Compared to the omni-directional solution, differential drive locomotion does not allow for instantaneous lateral motion. The remaining teams, iCub and Janus, exploited bipedal locomotion that can potentially provide the same flexibility of omni-directional wheeled systems by using moving limbs in contact with the environment. Nonetheless, only the iCub avatar managed to effectively use such a locomotion type during the scored trial.

One of the main challenges of bipedal locomotion is fall risk.  Walking is less stable, and asperities on the ground or accidental collisions may lead to a fall and damage to the avatar system. During the qualification day, the Janus avatar fell due to a malfunction, preventing the team from participating in the scored trials. During the first scored trial, the iCub avatar fell after hitting a pillar on the test course, hindering its possibilities to get in the top 12 positions allowed  for a second trial. Interestingly, during the Finals, the INBIODROID robot also fell, after a malfunction on the control of its wheeled base that caused a collision with one obstacle on the course.

Fall risk was a concern for other teams as well. Team SNU’s avatar is a bipedal robot, but for the sake of the competition it was placed on an omni-directional wheeled base. Similarly, team Avatar-Hubo’s avatar is a bipedal robot, but it stood on its legs only when still, while it used a set of wheels on its knees to move. Team i-Botics used a slender differential drive wheeled robot, and during its first scored trial, it was secured to an external gantry to prevent it from tipping over.

\subsubsection{Manipulation}

\begin{figure}
    \centering
    \includegraphics[width=0.96\linewidth]{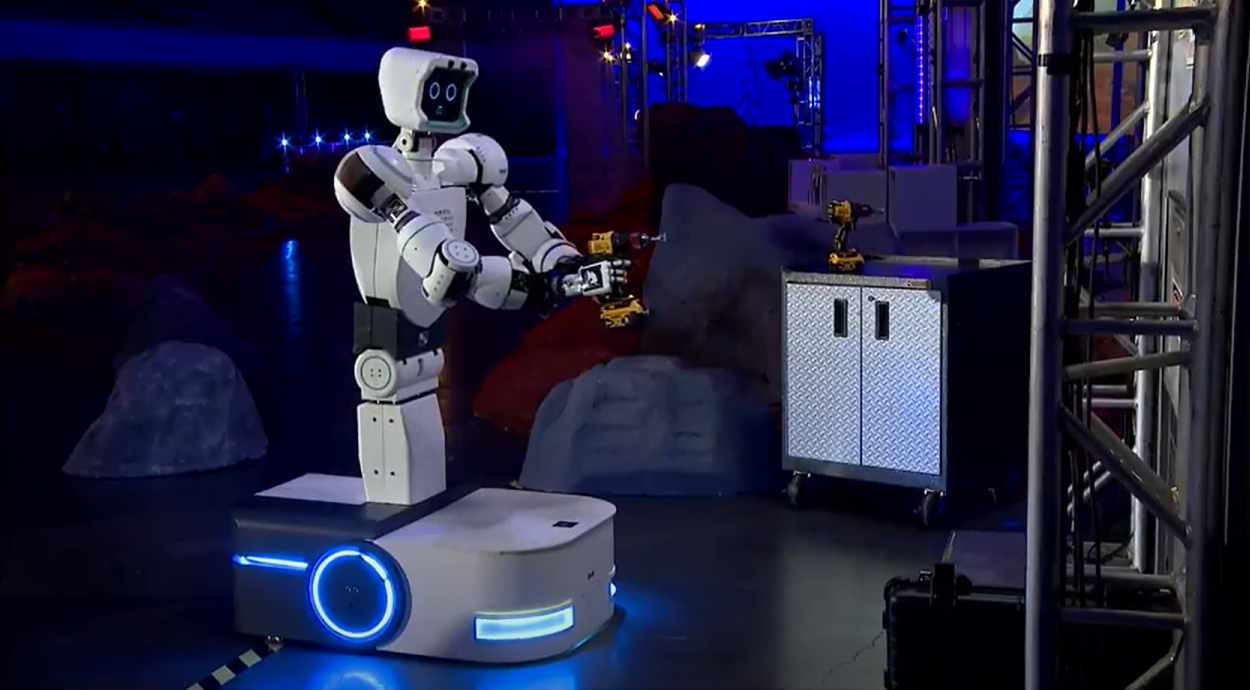}
    \includegraphics[width=0.96\linewidth]{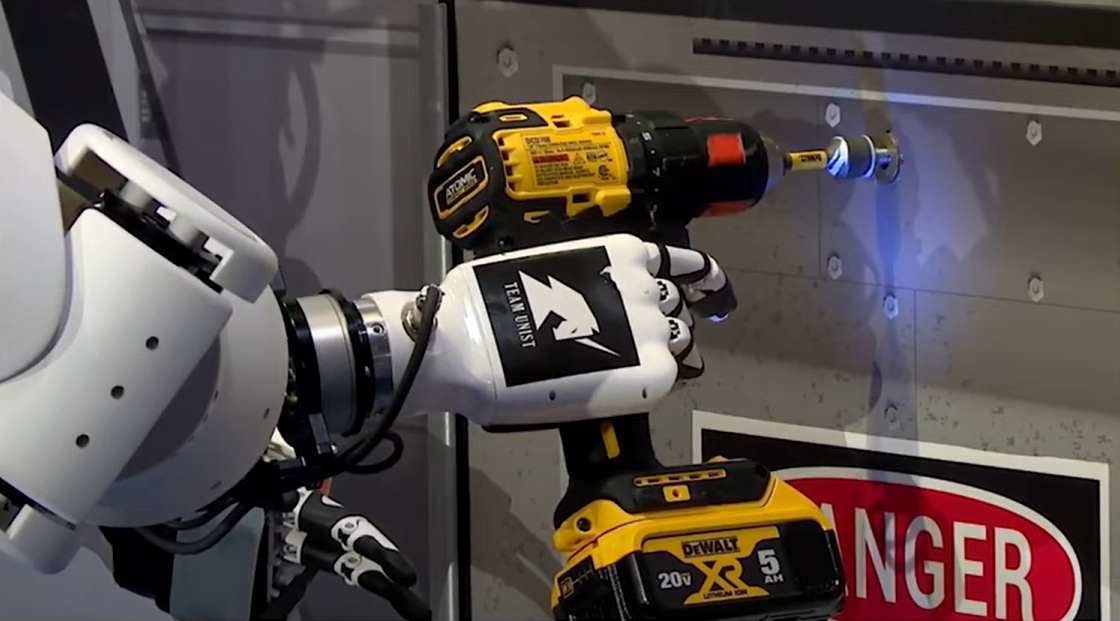}
    \caption{The drill task, as being attempted by Team UNIST.}
    \label{fig:FinalsSNUDrill}
\end{figure}

Providing the ability to manipulate objects through the avatars was a primary challenge of the competition. During the Finals competition, avatars were expected to lift up to 2.5~kg and have the precision and dexterity to operate a drill to unscrew an M12 bolt (Figure~\ref{fig:FinalsSNUDrill}).

Robot arms are used to control end effector motion and require 6 or more degrees of freedom (DoF) to achieve adequate position and orientation control with 0.5--1.5\,m reach and sub-centimeter precision. Teams adopted one of two approaches: either using commercially available {\em cobot} arms or developing custom actuators for the avatar.
Commercially available manipulator arms have been used extensively in factory automation and are known for their excellent precision, durability, and  integration readiness. Cobot (collaborative robot) arms are very similar in form except are certified to include safety features like collision detection and torque control that greatly reduce the risk of injury to humans that operate in the robot workspace. These arms are designed to be fixed on horizontal surfaces, but doing so for an avatar robot would appear unnatural. Hence, many teams have designed mounting positions and arm trajectories to look more human-like, mainly by modifying the base angle to better resemble a human shoulder and the elbow positions to resemble human posture. Using cobot arms helped teams assemble working solutions quickly, with 3 out of the 4 teams who completed all 10 tasks having chosen this option.

\begin{figure}[tbp]
    \centering
    \includegraphics[width=0.96\linewidth]{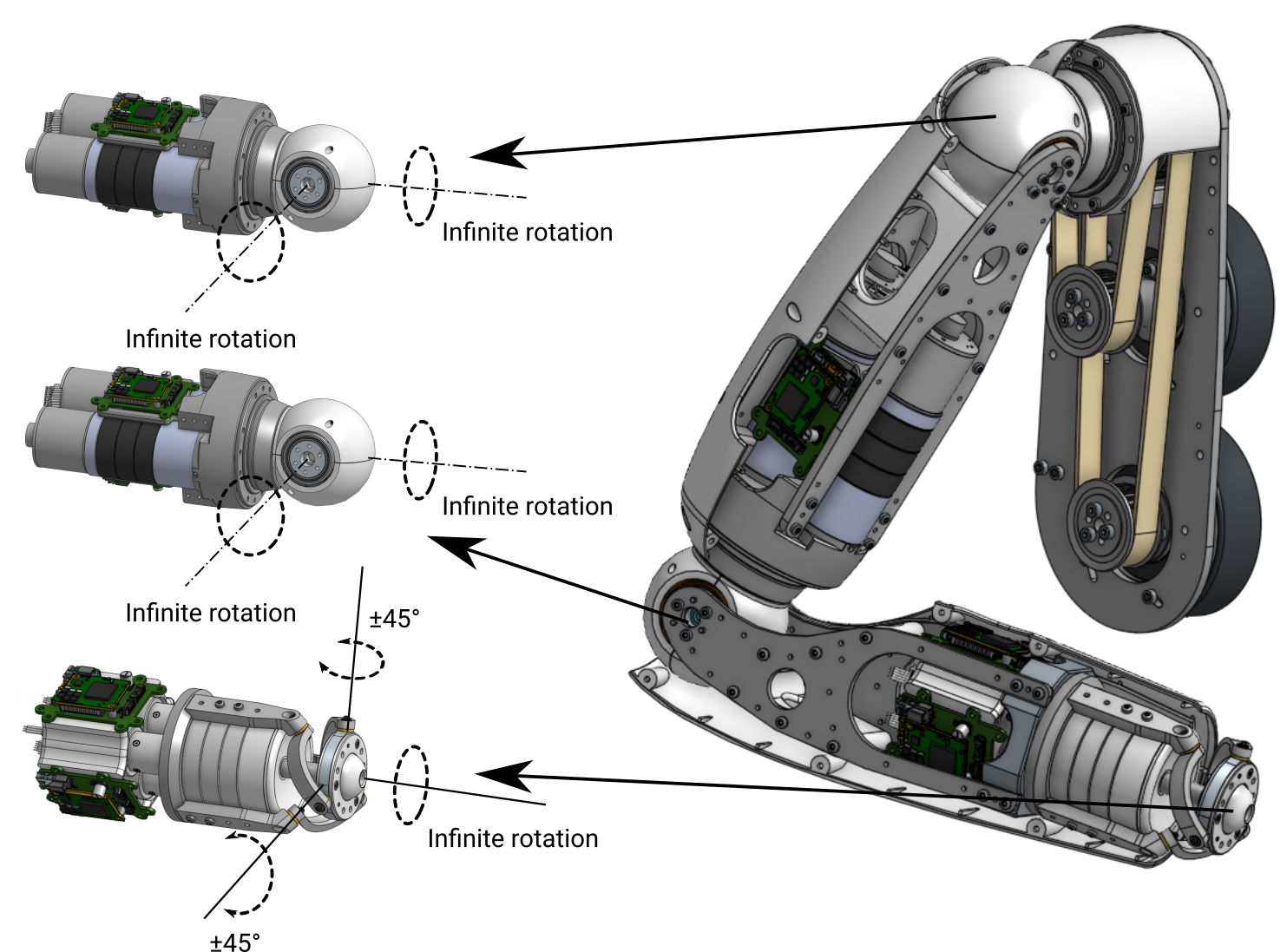}
    \caption{Team Pollen Robotics’ custom 7DoF arm, with two Orbita2D (shoulder and elbow) and one Orbita3D (wrist).}
    \label{fig:TeamPollenWrist}
\end{figure}

Some teams chose to develop custom actuators for their avatar's arms. Custom actuators could better  approximate human morphology for the avatars through bio-inspiration, but at the expense of greater engineering effort. Among the custom developed actuators, we highlight Pollen Robotics' Orbita joints, with a two-degrees-of-freedom and a three-degrees-of-freedom versions. 2D Orbita is a parallel spherical joint which has been used in the shoulder and in the elbow of their robot. This actuator combines several degrees of freedom like ball-and-socket joints in human shoulders and wrists, and permits continuous rotation. It allows for a more compact mechanism thanks to the combination of motors’ power and also enables the motors to be relocated. 3D Orbita has been used as a neck and wrist joint on the same robot. Its range is more limited on specific axes and increased on others, with infinite rotation around its vertical axis and 90$^\circ$ rotation amplitude (from -45$^\circ$ to +45$^\circ$) along the other axes  (Figure~\ref{fig:TeamPollenWrist}). Overall, this allowed the team to develop a lightweight and anthromorphic arm with sufficient force capabilities to complete the XPRIZE Finals.

Robotic hands are critical to teleoperation, especially for manipulation tasks. The finalists utilized different types of robotic hands to complete manipulation tasks, ranging from simple to complex. Robotic hands can be divided into parallel jaw and multi-finger types. A parallel jaw type refers to a structure in which the fingers are parallel and fixed, while a multi-finger type refers to a structure in which there are more than two fingers that are not parallel. Anthropomorphic design refers to a type of robotic hand with a structure that closely resembles the anatomy of a human hand. Of Finalist teams, 12 teams used anthropomorphic grippers, 2 teams used 3-finger grippers, and 5 teams used parallel grippers. This includes 3 teams that combined anthropomorphic and parallel grippers on different arms.  Typically, parallel jaw grippers are stronger, more precise, and less expensive than multi-finger grippers, whereas multiple fingers afford greater dexterity, flexibility in grip choice, and intuitiveness for operators.

Actuator transmission technologies include rigid, variable impedance via feedback control~\cite{hogan1985impedance}, and physical variable impedance~\cite{vanderborght2013variable,della2020soft}. Rigid actuation  implies a rigid connection between the motor, gearbox, and robot's link operated with high-gain position control, and characterizes classical industrial robot arms and servomotors used in grippers and other accessories. The risk of rigid control is that they are less safe than compliant control, particularly for robot arms which have strong motors. Variable impedance by feedback control is characterized by introducing torque sensors into the control loop to estimate and respond to load. This is characteristic of collaborative robots ({\em cobots}) used as the robot arms (Franka Emika Panda Arm, Universal Robots UR5) of several competition avatars. Physical variable impedance actuation is characterized by introducing compliant elements (e.g., springs) in the transmission, and was used by AlterEgo~\cite{lentini2019alter} in their arms. \revision{Several teams also employed underactuated grippers for their Avatars' hands, and in particular all anthropomorphic hands were underactuated using tendon or linkage-driven mechanisms. The number of actuated DoF vary per hand but each team's grippers used at least one DoF per finger (as opposed to a single hand synergy) due to the need to manipulate a drill trigger. }


\subsubsection{Vision and Head Movement} 

\begin{figure}
    \centering
    \includegraphics[height=.96\linewidth]{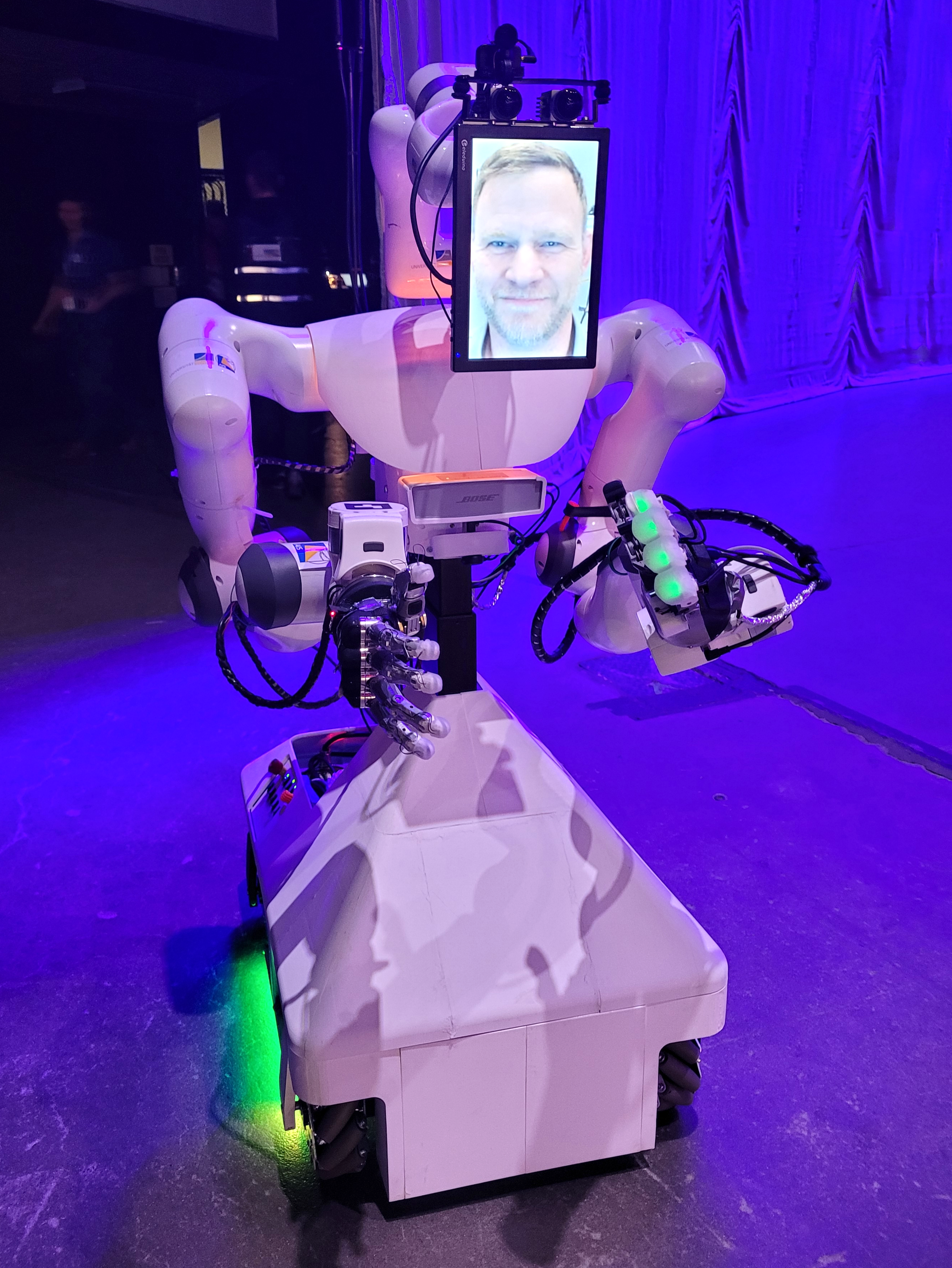}
    \caption{NimbRo~\cite{SchwarzNimbRo2022} avatar robot with anthropomorphic upper body and omnidirectional base. The robot's head is moved by a 6\,DoF arm and the operator's face is animated photorealistically.}
    \label{fig:NimbRoAvatarRobot}
\end{figure}

To provide immersive visual feedback to the operator, most avatar robots were equipped with a stereo camera system in their head. Resolution, field-of-view, and frame rate of these cameras varied. With higher-resolution high-frame rate cameras, image compression was mandatory to stay within the limits of the available WiFi bandwidth. A few teams used monocular cameras, which are unable to provide stereo depth perception to the operator, or RGB-D cameras, which tend to struggle obtaining depth readings on transparent or shiny objects.
Out of the 17 finalists, only 3 teams used a mono camera system, which were then rendered on standard screens on the HMI side. 

Some avatar robots were equipped with additional cameras to create a birds-eye view for navigation. Also horizontal LiDAR sensors were used to create a local map for obstacle avoidance and navigation assistance.
To aid manipulation tasks, some avatar robots were also equipped with in-hand or wrist cameras.

Most avatar heads were connected to the torso by a pan-tilt unit, tracking the viewing direction of the operator and extend their field-of-view in this way. The NimbRo~\cite{SchwarzIROS21} avatar robot moved its head by a 6\,DoF robot manipulator tracking the full operator head motion. This allowed for freely rotating the camera system and for 3D camera motion that permitted to chose a suitable viewpoint minimizing occlusion for manipulation tasks without robot base movement. The resulting motion parallax significantly contributed to the immersiveness of the visualization. Other neck-like mechanisms achieve fewer DoF movement, such as the pan-tilt-roll units in AVATRINA, iCub, Pollen, and UNIST.

Finally, most robots included a display showing the operator's face to contribute to the perception of their presence. Many teams  mounted the display on the same head unit that reflects the operator's head movement, which allows for more natural gestures to be communicated to humans in the remote environment, such as nodding the head.

\subsubsection{Force and Tactile Sensing}

Almost all robots in the finals were capable of force and tactile sensing due to competition requirements of sensing object weight and surface texture.  
To measure load on arms as typically caused by a grasped object, teams either used load-cells or joint torque sensors. Load cells may be built into a cobot's wrist (such as the UR5), mounted on the arm's wrist, or, less commonly, one of the other links. Joint torque sensors are available on many collaborative robots, which enables approximate reconstruction of the overall load using static analysis. 

All but one team in the Finals used some technique for surface texture sensing. Not only is touch sensation on the fingers and hands is an important component of human manipulation, it was a mandatory requirement to accomplish Task 10, which required blind reaching and identification of a textured object. Surface roughness generates high-frequency components of a haptic signal, and many types of sensors on the robot’s fingers and hands were used to measure roughness. Microphonic (NimbRo), accelerometer-based (e.g. AlterEgo), barometric (e.g. AVATRINA’s Takktile 2 sensors, SynTouch pressure sensors), electric-capacitive (SynTouch impedance sensors), and piezoresistive (eDermis e-skin used by Touchlab) are among the sensing technologies adopted. Almost all the teams mounted these sensors on the fingertips or on the palm of the end-effectors. 

Grasping force is another possible sensing modality which was only used by certain teams. Different technologies have been used including barometric (AVATRINA), current sensing at the actuation level (AlterEgo), hydraulics with pressure sensing (Team Northeastern), and piezoresistive (iCub). No teams used full-body sensing ``skin'' due to the complexity, brittleness, and relative immaturity of such technology.

It is important to note that all these force sensing technologies needed to be coupled with haptic devices at the operator station or alternative sensor rendering techniques, so teams carefully chose their haptic sensing technologies and dynamic range to correspond to the types of forces that could be adequately rendered.  As an example of a cross-sensory rendering technique, several teams used auditory cues to render surface texture.  As a result, full-body sensing ``skin'' would only be useful if a full-body haptic suit were worn by the operator, or some alternative augmented reality display were provided.

\subsection{Human Machine Interface Technologies}

The HMI technologies in the Operator System are meant to convey control commands from the Operator to the Avatar and sensory data from the Avatar to the Operator in an immersive fashion. \revision{Technologies used are  summarized in Table~\ref{tab:HMITable} and described in more detail below.}

\begin{table*}[tbp]
    \centering
    \includegraphics[width=0.96\linewidth]{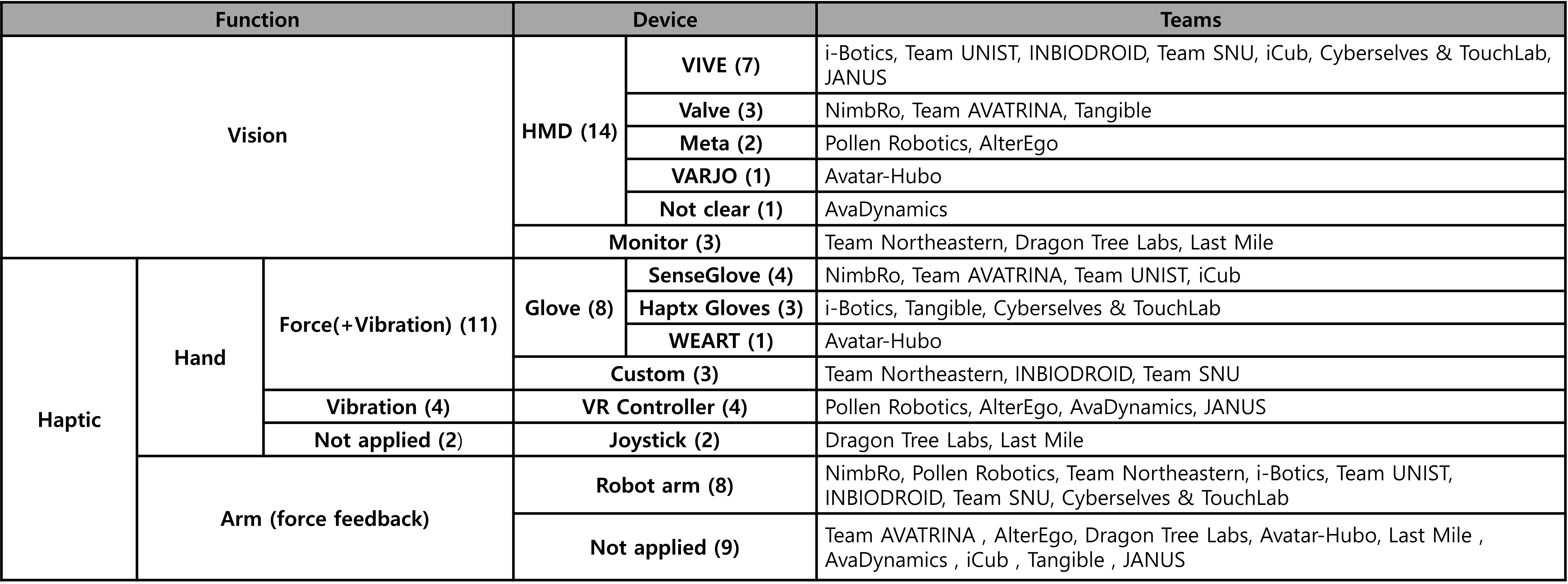}
    \caption{Breakdown of HMI technologies used by teams}
    \label{tab:HMITable}
\end{table*}

\subsubsection{Visual Displays}

Teams used two methods of presenting an avatar's vision to an operator: a head-mounted display (HMD) or an ordinary screen. HMDs have the advantages of a highly immersive experience and intuitive operation because the operator can change the direction of gaze by head movement. On the other hand, it requires time and effort to put on the HMD, and if the change of the visual field image is not consistent with the head movement, it may cause motion sickness. Screen displays  provide a familiar ``computer console'' experience. Advantages of this method include lower time and effort to don an HMD, improved comfort, and elimination of motion sickness.

Since one of the objectives of this competition was to construct a highly immersive teleoperation system, 14 of the 17 teams adopted  HMDs.  HTC VIVE was the most popular HMD, while HMDs from Valve and Meta were also employed.  In order to suppress motion sickness when using an HMD, methods such as changing the position of the visual field image displayed in the HMD according to the displacement~\cite{Chen2022} and generating a visual field image corresponding to the viewpoint using spherical rendering~\cite{Schwarz2021} were used. Three teams (Team Northeastern, Dragon Tree Labs, and Last Mile) used static screens, but to enhance the immersive experience, most teams used a large and curved display to show what was in front of the robot in full scale. These could be mounted either vertically (Team Northeastern, Dragon Tree) or horizontally (Last Mile),

In terms of rendering the operator's presence to the remote environment, one issue with the use of HMDs was the occlusion of the operator's face. This hindered the direct capture and display of facial expressions. 
To address this issue, some teams used HMD-mounted cameras to capture the mouth region and the eye movements inside the HMD. NimbRo developed a method to estimate gaze direction and eye opening state as well as facial feature points for the mouth region and trained a morphing network to animate the operator face in a photorealistic way~\cite{RochowIROS2022}, see Fig.~\ref{fig:NimbRoAvatarRobot}. Other teams (AVATRINA, Pollen) used similar techniques for face rendering.

\subsubsection{Manipulation and Navigation Control} 

Avatar arm movement was typically accomplished using human-robot motion retargeting. Operator movements are recorded using one of two primary tracking modalities: grounded exoskeletons or visual motion capture.  Many high-scoring teams (NimbRo, Team Northeastern, iBotics, Inbiodroid) utilized the former as it provided a direct avenue to implement arm force feedback.  Contrarily, most (Avatar-Hubo, iCub, Pollen Robotics, Tangible Robotics, AVATRINA) employed the latter since many reliable, safe, and comfortable off-the-shelf VR motion-capture devices (notably the HTC VIVE Trackers and SteamVR Base stations) are readily available.  While wearable motion-capture technology is less restrictive than its alternative, lacking a straightforward way to implement operator force feedback required additional consideration of avatar arm-environment contact forces. 

Gripper control was achieved using either buttons on VR controllers or glove-based devices. Many teams used off-the-shelf gloves, such as the SenseGlove or HaptX, that read finger positions and provide force and vibrotactile feedback. Usually, teams that used multi-finger and anthropomorphic grippers on their Avatar system opted for the use of gloves, which provide more natural 1-to-1 motion control.  Notable exceptions include Pollen Robotics and AlterEgo, which opened and closed 3-finger and 5-finger grippers (respectively) using a single VR controller button. Compliance around a target object during the closing motion was achieved using an underactuated mechanism. Parallel jaw mechanisms can be controlled more easily through buttons or joysticks.

\begin{figure}
    \centering
     \resizebox{\columnwidth}{!}{
    \begin{tikzpicture}
    \begin{axis}[
        axis y line       = left,
        axis x line       = bottom,
        tickwidth         = 5pt,
        enlarge y limits  = 0.2,
        xmin              = 0,
        xmax              = 5,
        width             = 1.5\columnwidth,
        height            =0.8\textheight,
        yticklabels={
            Foot Motion\\Controller,
            Handheld VR\\Controller,
            Desk\\Joystick,
            Pedals,
            Sensorized\\Shoes,
            Gaming Console\\Controller
        },
        ytick             = data,
        xtick             = data,
        bar width         = 45pt,
        xlabel            = Number of Teams,
        xlabel style={
        font =\large,
        align=center
        },
        major grid style  = {
            line width = 0.2pt, 
            draw=gray!50
            },
        ymajorgrids=false,
        xmajorgrids=true,
        xminorgrids=false,
        major tick length=3mm,
        minor tick length=2mm,
        y tick label style={
        font =\large,
        align=center
        },
        x tick label style={
        font =\large,
        align=center
        },
    ]
    \addplot[xbar,fill=blue] coordinates {
        (5,5)
        (4,4)
        (3,3)
        (2,2)
        (2,1)
        (1,0)
    };
    \end{axis}
    \node[font=\small, text=white, align=center] at (5,13.5) {Cyberselves $\vert$ Touchlab, i-Botics,\\ NimbRo, Team Northeastern, Team UNIST};
    \node[font=\small, text=white, align=center] at (4,11.3) {AlterEgo, AvaDynamics, AVATRINA,\\ Pollen Robotics};
    \node[font=\small, text=white, align=center] at (3,9) {Dragon Tree Labs,\\ Last Mile, Tangible};
    \node[font=\small, text=black, align=center] at (6.1,6.7) {INBIODROID, Team SNU};
    \node[font=\small, text=black, align=center] at (5,4.5) {iCub, Janus};
    \node[font=\small, text=black, align=center] at (3.1,2.2) {Avatar-Hubo};
    \end{tikzpicture}
    }
    \caption{Devices used to control the avatars’ locomotion.}
    \label{fig:NavigationControl}
\end{figure}
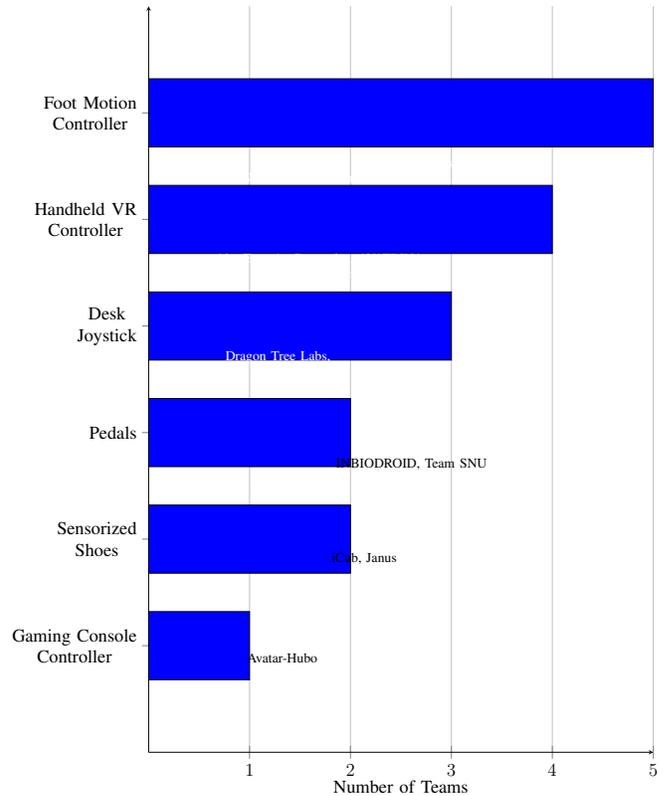

Teams implemented different strategies to control  locomotion. A summary of the adopted devices is in Figure~\ref{fig:NavigationControl}
The choice of input device for the locomotion control depends on different factors, and teams chose a variety of handheld and foot-operated devices.

Hand-operated devices include handheld VR controllers, desk joysticks, or gaming console controllers. The use of a handheld VR controller limits the possible manipulation commands that the operator can issue with the same hand, but most controllers have sufficient numbers of buttons to operate parallel-jaw grippers.  A multi-modal approach taken by Dragon Tree Labs and Last Mile allows switching between locomotion and navigation using a toggle button. Similarly, Tangible and Avatar-Hubo adopted a user interface in VR to switch control modes.  Desk joysticks and gaming console controllers were used by teams with standard screen displays, so device switching is more convenient than in VR. 

Foot-operated devices free up the hands to perform simultaneous manipulation and navigation control. Six teams have used either commercial (e.g. 3dRudder\footnote{https://www.3drudder.com} and HoboLoco\footnote{https://www.hoboloco.com}) or custom devices to measure foot motion along 3 axes, or a combination of pedals, thus controlling directly the motion of the wheeled base. Team iCub instead used a pair of iFeel\footnote{https://ifeeltech.eu} sensorized shoes to detect the operator taking footsteps in place and control the iCub avatar stepping sequence accordingly.

\subsubsection{Force Feedback}

Most teams used some form of vibrotactile transducers inside their manipulation controllers (VR controller or glove) to render oscillatory forces or cues to the operator (eccentric motor, electromagnetic, or piezoelectric).  These were primarily used for texture identification (Task 10), and several teams also augmented vibration with an audio signal to provide a stronger sense of texture. Vibrotactile transducers are inexpensive and could be placed at multiple locations on the body~\cite{lieberman2007tikl}, but the XPRIZE Finals guidelines did not require rendering bodily sensations so teams did not explore such configurations further.

To render static grasp forces on fingers many teams used ``brake-type'' actuators/clutches to render uni-directional grasp forces, such as the SenseGlove DK1 (magnetic friction brakes) and HaptX (tendon brakes).  Team Northeastern used a classic force-reflecting haptic interface arrangement for three finger DoFs to provide combined DC and AC force rendering.  Finally, localization of contact was provided by HaptX microfluidic surface actuators, used by two teams.

To provide haptic feedback to the hands, 11 teams used gloves, of which 9 teams used SenseGlove, HaptX Gloves, or WEART with force and vibration feedback. 2 teams utilized VR controllers for vibration feedback, while 2 others used joysticks. 

8 teams utilized either end-effector type exoskeletons or robot arms to provide force feedback to the arm. Out of these teams, NimbRo, Cyberselves \& TouchLab, and i-Botics used commercially available robot arms or haptic devices such as Franka Emika, Universal robots, and Virtuose, respectively. The remaining teams employed their own end-effector type exoskeletons, which provided 6\,DoFs or 3\,DoFs force feedback at the hand, or 1\,DoF force feedback at the elbow (Pollen Robotics). The majority of the teams that incorporated force feedback devices for the arm outperformed those without such devices.

\subsubsection{Augmented Reality and Shared Control}

Immersive telepresence typically calls for direct control of the avatar robots based on visual and haptic feedback. Nevertheless, many teams developed augmented reality (AR) display components for operator assistance. These include, for example, smart watch displays on the wrists of the NimbRo robot arms showing time and the measured weight of the grasped object. Another example are laser lines projected by the Northeastern avatar to compensate for the lack of depth perception in navigation and manipulation. Avatar models were visualized to help with initial alignment of the operator arm poses with the avatar arm poses~\cite{LenzRAS2023} and to indicate base movement predictions for anticipatory navigation. 
One team (NimbRo) even showed a ``cheat sheet'' to help the operator remembering how to solve the challenge tasks.

Additional sensors could also be included to augment the primary head point of view, including birds-eye camera views (Northeastern, NimbRo) or horizontal LiDAR scans (Pollen) to aid navigation. 
In-hand cameras and depth sensors were used to aid manipulation in many teams, especially for Task~10, where the direct view from the robot head was blocked. AVATRINA used depth sensing to produce an AR heightmap and ``virtual stylus'' to allow the operator to feel the texture of the rock without contacting it with the robot.

Shared control policies were adopted by all teams with legged or self-balancing avatars for the balance controllers, which were mixed with the navigation intentions of the operator (e.g. iCub, i-Botics, AlterEgo). For manipulation control, only a few teams used a shared autonomy approach.  Only four finalist teams (AlterEgo, Avatar Hubo, LastMile, and AVATRINA) declared the use of operator assistance functions, e.g. to control redundant DoFs of the avatar arms (elbow), or to generate repelling forces from obstacles or joint limits~\cite{LenzRAS2023}. It is likely that the scope of the competition dissuaded teams from investing effort in the development of shared autonomy tools, except for those strictly necessary to operate the robot. 

\subsubsection{Communication Technologies}

The quality of recipient interfaces (e.g. audio clarity, depictions of the operator's facial features, emotions, and gestures) integrated into Avatars influence the ability of operators to communicate effectively to recipients. 

Audio capture and display were relatively straightforward for this competition; standard microphones and speakers were sufficient for teams to adequately convey speech communication. We do note that robot and computing machinery may emit noises that disrupt audio communication, and hence microphones and speakers should be suitably insulated or placed away from sources of noise on the robot.  

Several teams provided a facial display of the operator on the robot as a recipient interface.  Solutions ranged from rendering static facial iconography (iCub, Team SNU), to animated anthropomorphism (Avatar-Hubo, INBIODROID, Team UNIST), or even dynamic reconstruction of an operator's actual face (NimbRo, AVATRINA, AvaDynamics, Tangible Robotics).  The differences in both recipient interface strategies and Mission Commanders complicated direct comparison between teams; for instance, both animated and realistic representations of the operator's face could clearly exhibit facial features, emotions, and gestures.  Nonetheless, increased facial emotive dynamics can convey greater expression than static faces, possibly easing initial interactions with the operator through their Avatar.

Nonverbal communication was explored by operator judges to help facilitate communication. Several operators  leveraged head, arm, and base motion to convey gestures and body language. These efforts aided in developing the social setting between the operator and recipient judges, especially since the latter was prevented from reacting to the operator until they introduced themselves.  Notably, a recipient stated that this process quickly demonstrated the Avatar's dexterity, nimbleness, and interaction capabilities nonverbally. Operators also notably performed celebratory gestures such as arm waves or spins to convey positive emotional state after completing tasks.

\subsection{Software and Networking Infrastructure}

Networking infrastructure was a critical component of the competition and this required the teams to take into account the latency and throughput of communications between the operator and their avatars. During the Finals, XPRIZE provided a 2.4\,GHz and a 5\,GHz WiFi network in the competition arena. Teams recorded latency values of 1~ms on average with worst-case of \~100~ms. Low latency was particularly important for haptic feedback. To this end, some teams employed UDP-based network transport solutions\footnote{NimbRo\_network: \url{https://github.com/AIS-Bonn/nimbro_network}} \footnote{UDPROS: \url{http://wiki.ros.org/ROS/UDPROS}} instead of TCP. The Network Device Interface (NDI)~\cite{NDI} and WebRTC~\cite{WebRTC} protocols for the transmission of video and audio signals were utilized by a number of teams. The bandwidth for video transmission was variable to adapt to different network conditions and the resolution and bitrate of the videos have been adjusted during the competition. The network latency presented a significant challenge in loudspeaker systems, causing significant acoustic feedback. To mitigate this issue, active echo cancellation (AEC)~\cite{deb2014technical} was commonly used. As an alternative, some teams opted to suppress the microphone while the speaker is in use during audio-visual communications. For low-latency high-fidelity audio-visual operator feedback, it was imperative to compress the data streams with codecs that used small buffers. It was also necessary to constantly monitor WiFi health and to adapt the transmitted data accordingly. Some teams even used redundant transmission over both WiFi frequencies.

As teams were down-selected on each of the three Finals days and the tasks were to be performed in the given order, a single failure could result in not advancing. Consequently, system reliability was of utmost importance and some teams developed system health monitors and automatic recovery procedures to continue even in case components should fail~\cite{SchwarzNimbRo2022}.

\section{Results and Analysis}
\label{sec:Analysis}
\subsection{Finals Competition Results}

The official final ranking is shown in Table~\ref{tab:FinalScoring}. Overall, 28 runs were conducted during the two days of Finals, with 12 teams able to complete two runs.  The entire 10 task course was completed by 4 teams on 6 runs, with 2 teams completing the entire course on both their allotted runs. 

\begin{figure}[tbp]
    \centering
    \includegraphics[width=0.96\linewidth]{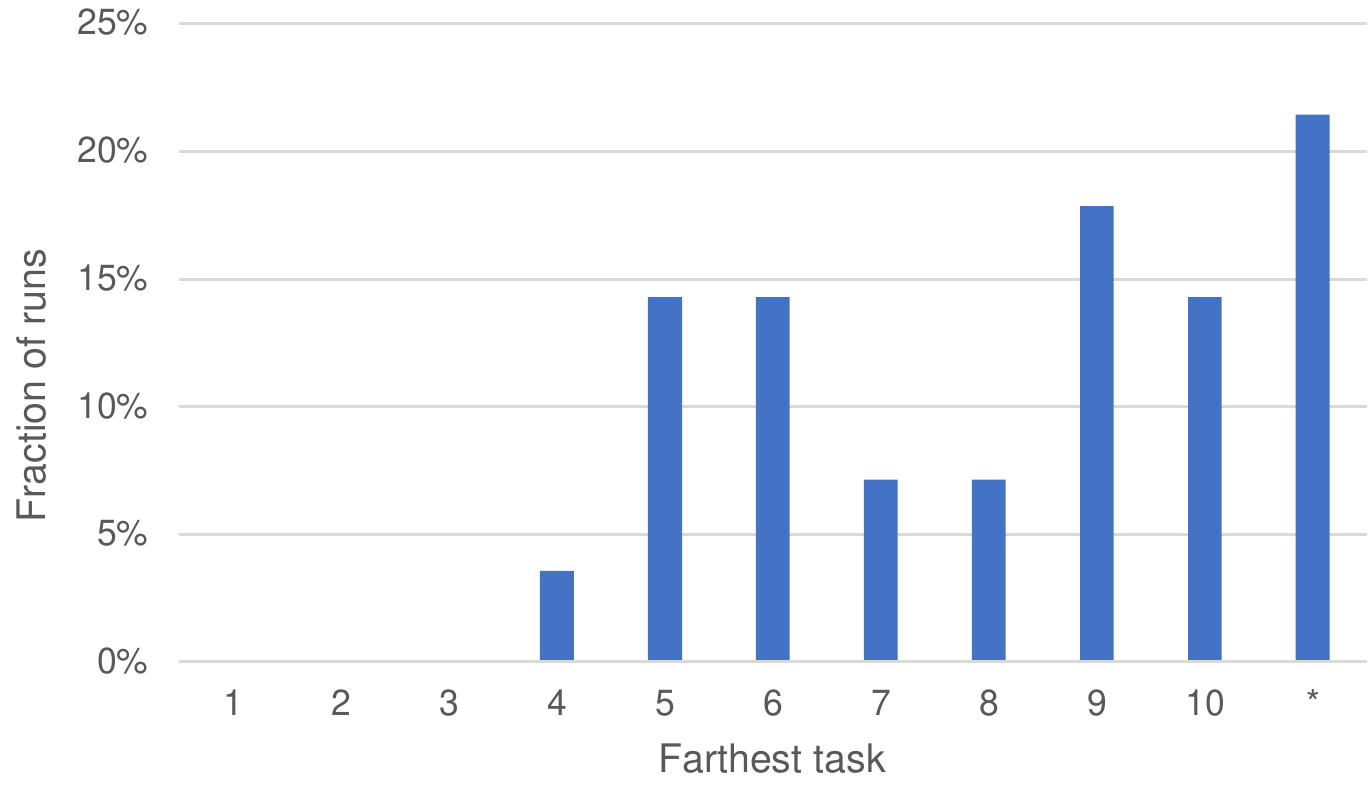} 
    \caption{Frequency of farthest task encountered during a run  (i.e., the task being attempted at which a run failed) at Finals. * indicates the course was completed. The Drill task (Task 9) was the most challenging while the Mission Communication (Tasks 2--3), switch (Task 4), navigation (Tasks 1 and 8), and Canister Plug-In (Task 7) were the least challenging.}
    \label{fig:FinalsFarthestTask}
\end{figure}

Figure~\ref{fig:FinalsFarthestTask} plots the breakdown of final task for each run. Task 9, the drill task, was the most challenging, while the navigation (Task 5), canister lifting (Task 6), and rock identification (Task 10) tasks also had considerable drop-out. No teams failed Tasks 1--3, which could be accomplished with any device with mobility and A/V communication, such as a standard telepresence mobile robot.  Task 5, long-distance navigation, was failed twice because teams had delays starting their robot or trouble completing Task 4, leading to a timeout during this task.  Two other teams failed by colliding with the gate before the traversal.  The drill task was a challenge for many teams due to the need to accurately position a finger on the drill trigger, lift the relatively heavy drill, and precisely align the drill driver with the bolt. Two drills were provided in case one was dropped, which occurred in many runs. 

\begin{table}[tbp]
    \centering
    \includegraphics[width=\linewidth]{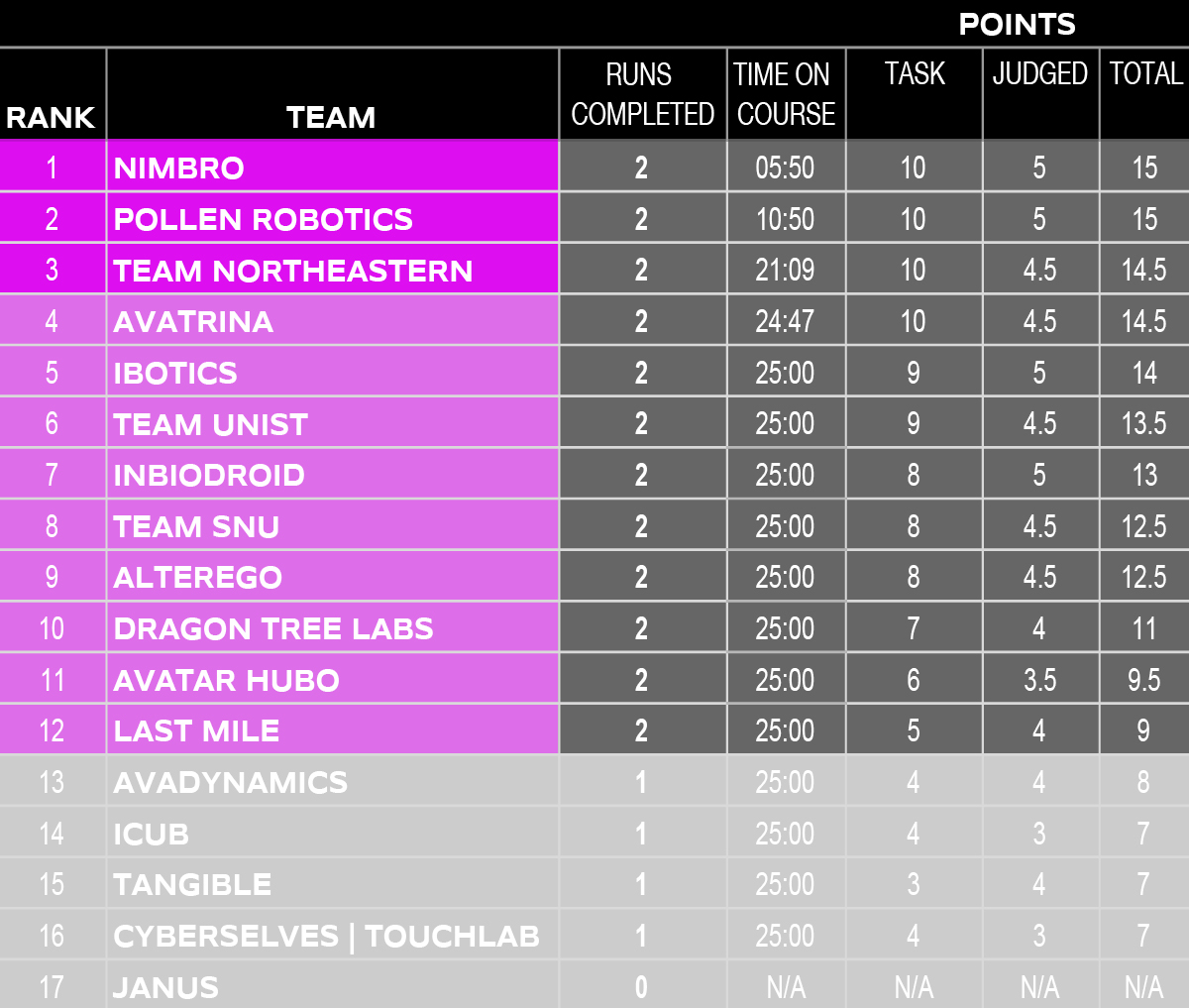}
    \caption{Finals competition ranking (best score of two runs, if applicable)}
    \label{tab:FinalScoring}
\end{table}

\begin{figure}[tbp]
    \centering
    \includegraphics[width=0.98\linewidth]{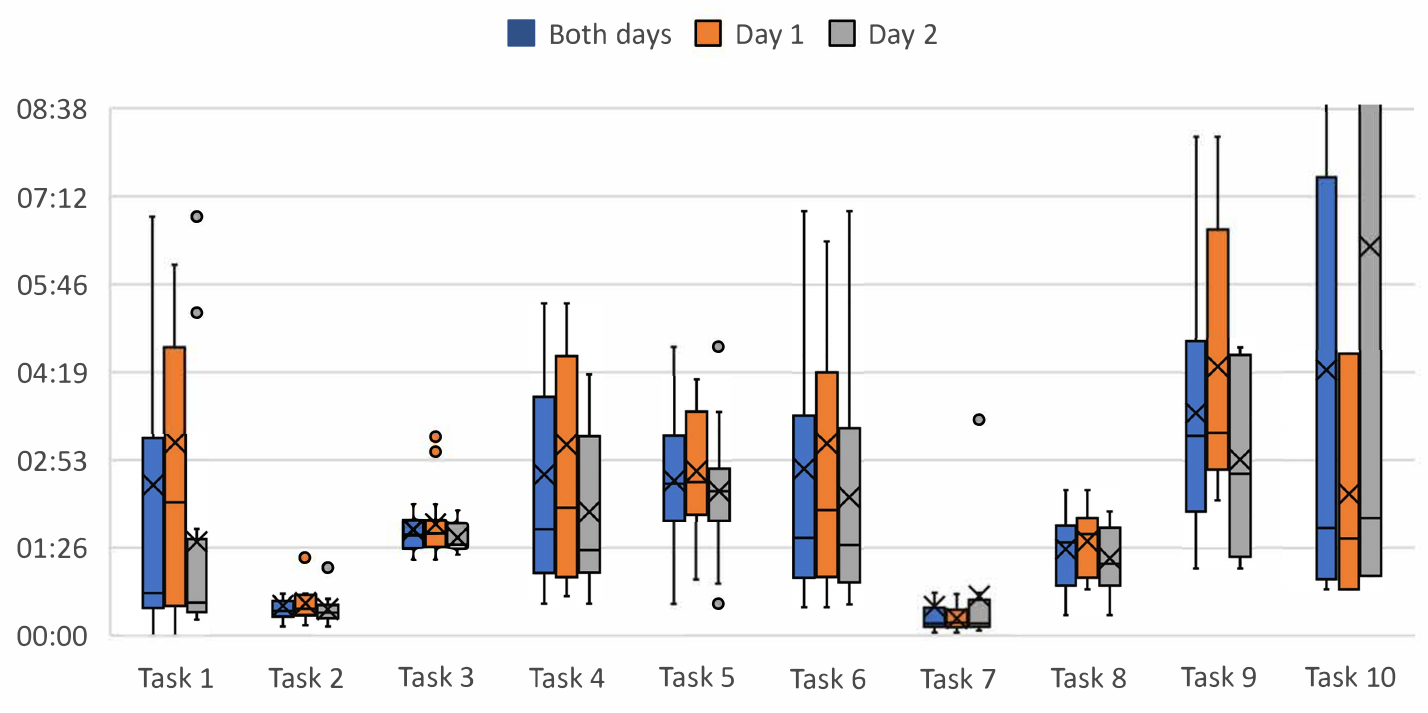} 
    \caption{Time to complete each task during Finals (minutes:seconds). Boxes denote lower and upper quartiles, lines denote median, and crosses denote mean. Task 1 and Task 9 showed major improvements from Day 1 to Day 2. }
    \label{fig:FinalsTiming}
\end{figure}

Timing was a considerable factor to determine the ranking of top teams.  NimbRo's second run was the fastest of the competition, completing the course in less than 6 minutes, while Pollen Robotics' second place time was approximately 11 minutes. Figure~\ref{fig:FinalsTiming} breaks down the timing between tasks. Much of the variation of Task 1 is related to network connectivity or system setup problems after the start of the run; once the Avatar started moving the vast majority of teams completed Task 1 within 45\,s.  Very few teams also completed Task 10, leading to significant variation due to small sample size. Overall, times tended to show improvement from Day 1 to Day 2. This can be explained as Judges gained experience with operating robots, and perhaps attempted to complete the course more aggressively on behalf of teams.  

\begin{figure}[tbp]
    \centering
    \includegraphics[width=0.96\linewidth]{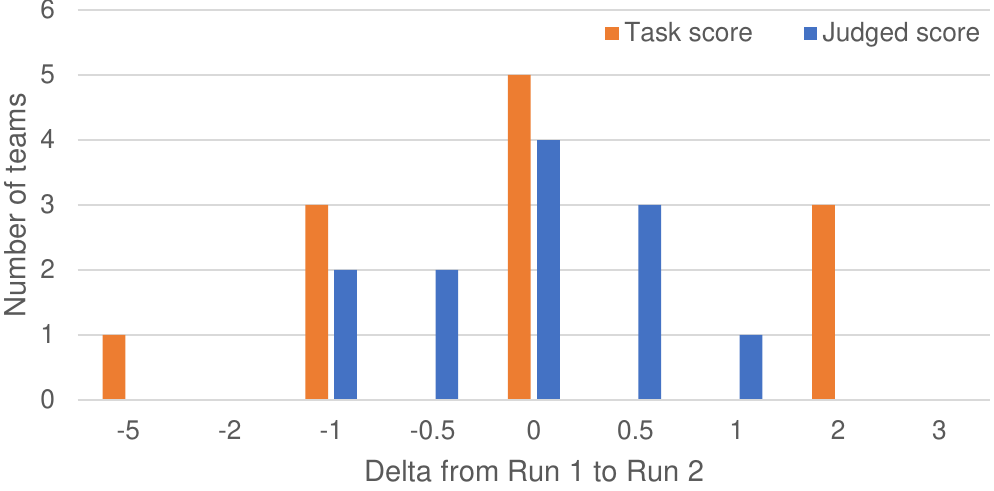} 
    \caption{Histogram of score deltas between Run 1 and Run 2 at Finals amongst teams that attempted two runs. Positive values indicate point total increases.  Scores between runs were fairly consistent, especially for judges' subjective scores. }
    \label{fig:FinalsDeltas}
\end{figure}

Teams failed to complete the course for a number of reasons, including running out of time, hardware breakage, emergency stop triggers, and irrecoverable task failure (such as dropping the two provided drills).
It was noted by the Judging panel that a few teams were reluctant to engage the E-Stop to protect their robot from damage. This was likely influenced by the fact that if the robot was E-Stopped the run was over.  As a result, some teams suffered catastrophic hardware failures that prevented a second attempt.

\subsubsection{Competition Scoring: Validity and Reliability}

Assessing the validity and reliability of any competition necessitates deep analysis, and the XPRIZE Finals raises several significant concerns: teams were only tested twice, with subjective metrics, with course conditions never experienced until a day before the competition, and with high-stakes methodology in which task failures end a run.  

Reliability of the scoring metric can be assessed by examining run-to-run differences. Figure~\ref{fig:FinalsDeltas} shows a histogram of differences between Task scores and Judged scores between runs. We see that Judged scores exhibited much smaller variance than Task scores, with most differences falling within 0.5 points. A few teams increased Task score by +2 points, which usually was achieved by better operator training, hardware changes, or software tweaks after the first run. One team lost 5 points because of unexpected software or networking failure in the early stages of their second run.

\subsection{From Semifinals to Finals}
\label{sec:SemifinalsToFinals}

\begin{figure}[tbp]
    \centering
    \includegraphics[width=0.96\linewidth]{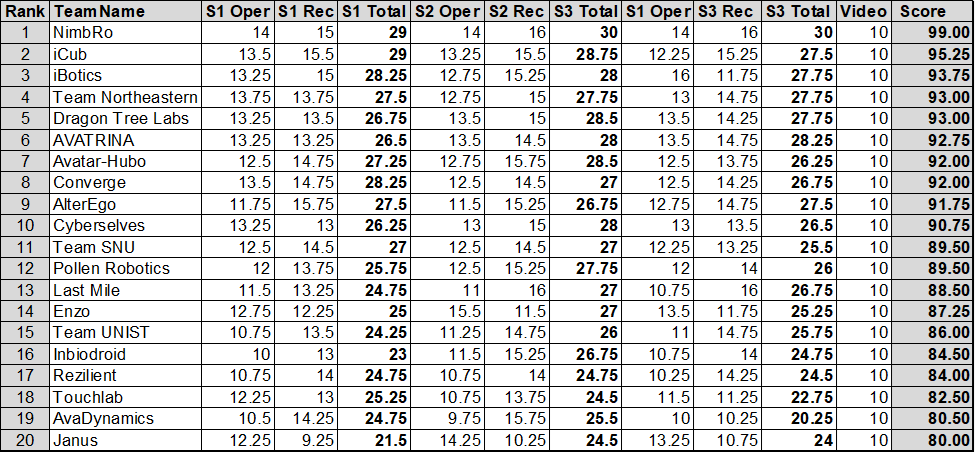}
    
    \vspace{0.2cm}
    \includegraphics[width=0.96\linewidth]{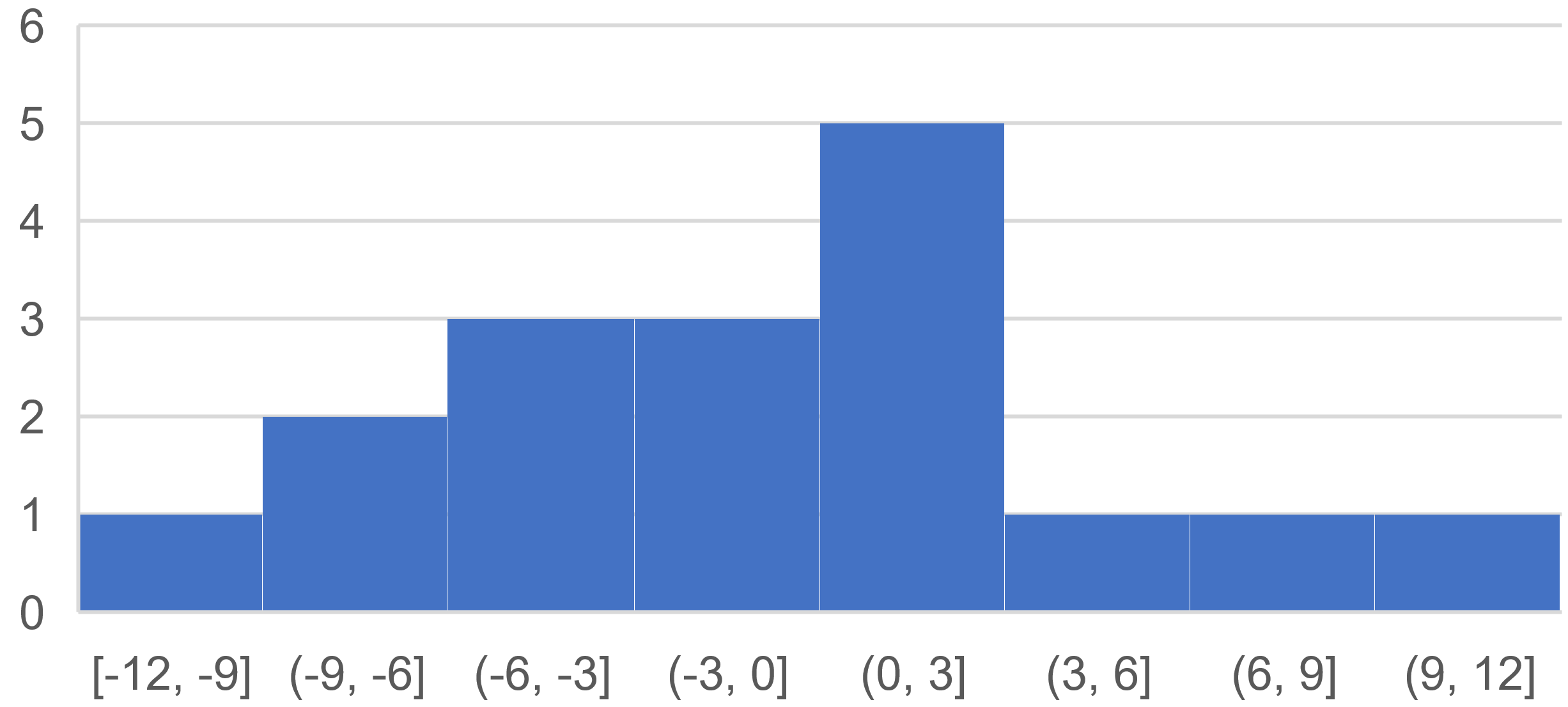}
    \caption{Ranking at Semifinals competition for Finalist teams (best of two runs) and a histogram of ranking deltas from Semifinals to Finals (positive indicates a better ranking at Finals). The largest ranking changes were Pollen Robotics (12$\rightarrow$2) and iCub (2$\rightarrow$14). }
    \label{fig:SemifinalScoring}
\end{figure}

Semifinals scores for finalists and the deltas between Semifinals and Finals scores are shown in Fig.~\ref{fig:SemifinalScoring}.  Many teams changed rankings drastically between these phases, which can be explained by differences in the emphasis of tasks, scoring of subjective vs objective components, and networking troubles.

At Finals testing, XPRIZE introduced some notable changes to the evaluation environment. First, the Avatar robot was required to run untethered and rely on battery power and wireless communications. Second, the Operator System was required to be moved into the Operator Room approximately one hour before testing time, rather than remaining resident as in Semifinals. Third, the course required the Avatar to complete much more significant navigation and fine manipulation tasks. Fourth, the format of the course was changed to a linear sequence of high-stakes tasks, which disincentivized teams from taking a holistic, generalist approach to developing their systems.  Teams had 6 months between specification of the Finals tasks and Finals competition, and the requirement to adapt to these changes favored larger, well-funded teams willing to fine-tune their system to the Finals course. 

We observe that 66\% of the scoring point totals were subjective in Semifinals testing compared to 33\% at Finals.  Based on the original statement of the XPRIZE (Sec.~\ref{sec:OrganizationBackground}) human experience was originally deemed by the Judging Panel a priority for evaluation.  However, after reviewing the results of Semifinals testing, XPRIZE observed a great deal of convergence amongst top teams in human-human interaction factors. This is largely due to technological consistency in video and audio communication methods which were found to be adequate for most forms of non-physical interaction. For example, 96\% and 99\% of the Semifinals entries satisfied the Recipient rubrics ``I was able to understand the Operator's communications'' and ``I felt the Operator could understand me'', respectively.  Moreover, 92\% of entries satisfied the Operator rubric ``I was able to sense the Recipient's emotion.''

Hence, changes were made to the Judging scoring process for Finals. First, the number of subjective questions was reduced from 20 to 5. After analyzing the scores of teams advancing to Finals it was clear that many of the Experience objectives of the program were realized by the teams and evaluating them again wouldn't provide separation in Finals evaluations. Second, the number of Operator Judges was reduced to use fewer Judges operating more teams. This pivot was based on observing that Judges who operated more often tended to be more consistent in their scoring. Third, the gradient of the scoring scale for objective questions was reduced from 5 options to 3. The gradient of scoring at Semifinals was interpreted  differently between judges, leading to less consistent scoring than at Finals. Lastly, XPRIZE selected the Judges with the most experience in Semifinals for Operators at Finals. This worked to ensure all teams received Judges with multiple runs of experience.

It should also be noted that several teams had issues with the network at Finals, and amongst the teams most severely affected were those unable to participate in Semifinals Plan 1 (onsite) testing. The Plan 2 teams (AlterEgo, iCub, and i-Botics) were not required to test on the Avatar Network at Semifinals and had less time to anticipate network conditions at Finals.

\subsection{Operator Judge Survey}

Operator Judges were asked to complete a 23-question survey about their impressions of each Avatar system after each run during Finals testing. The survey questions were proposed by a research team at University of Illinois (coauthors of the paper) and responses were gathered by XPRIZE.  Table~\ref{tab:SurveyCorrelations} lists all of the survey questions. Note that these are listed descending order of correlation to the Operator Score, not the order in which they were administered.  Each question was rated on a 5-point Likert scale with ratings 1 indicating ``worst'' and 5 indicating ``best'' for some questions and the reverse polarity for other questions.  We also collected an {\em Overall Rating} which indicates the sum of survey scores, corrected for questions with negative polarity (higher = better).

Standard deviations amongst all ratings were between 1 and 1.6. Inter-judge consistency of ratings of the same team (i.e., between runs) was between 0.9 and 1.6 RMSE for all questions except for sound localization (Q13) and quick learning (Q20), which were higher.   We observe that manipulation proficiency (Q14), feelings of immersion (Q3, Q12), visual fidelity (Q10, Q6, Q5), and feelings of proficiency  (Q9) correlated highly with both Operator Scores and Task Scores.  Surprisingly, latency (Q7) and ease of training (Q18, Q20) were not amongst the factors most correlated with operator or task score. We note that the top two teams were rated as low latency, and most teams were rated to have high ease of training.  

\revision{These results indicate moderate consistency between judges' subjective ratings, with magnitudes in line with the consistency in subjective scoring from Day 1 to Day 2 in Figure~\ref{fig:FinalsDeltas}. Interestingly, we find that amongst the top four teams, inter-judge ratings are significantly more consistent (45.7\% lower RMSE) than the remaining teams.  This suggests that high-performing teams may have designed their interaction paradigms to provide more consistent interpretability. Another possible explanation is variable attitudes toward how harshly to criticise low-performing teams.}

\begin{table*}[tpb]
\footnotesize
    \centering
    \renewcommand{\arraystretch}{1.2}
    \begin{tabular}{@{}p{0.4cm}p{12.5cm}p{0.8cm}p{0.8cm}@{}}
\toprule
{\bf \#} & {\bf Question} & {\bf Oper. Score} & {\bf Task Score} \\
\midrule
14& How well could you manipulate objects in the remote environment? & 0.81 & 0.63 \\
3& How much did your experiences in the remote environment seem consistent with your real world experiences? & 0.67 & 0.60 \\
10& How much did the visual display quality interfere or distract you from performing test course tasks? [-] & -0.65 & -0.47 \\
  & {\em Overall Rating} & 0.65 & 0.52 \\
6& How closely were you able to examine objects? & 0.64 & 0.36 \\
5& Were you able to survey or search the environment using vision? & 0.63 & 0.49 \\
9& How proficient in moving and interacting with the remote environment did you feel at the end of the experience? & 0.62 & 0.43 \\
12& How well could you concentrate on the test course tasks rather than on the mechanisms used to perform those tasks? & 0.60 & 0.64 \\
15& I would be comfortable using this system frequently. & 0.58 & 0.49 \\
8& How quickly did you adjust to the remote environment experience? & 0.57 & 0.47 \\
4& Were you able to anticipate what would happen next in response to the actions that you performed? & 0.56 & 0.51 \\
1& How much of the system were you able to control? & 0.55 & 0.39 \\
19& I found the functions in this system were well integrated. & 0.48 & 0.51 \\
22& I felt confident using the system. & 0.48 & 0.34 \\
17& I thought the system was easy to use. & 0.47 & 0.40 \\
21& I found the system very cumbersome to use. [-] & -0.43 & -0.33 \\
2& How natural was the mechanism which controlled movement through the test course? & 0.39 & 0.25 \\
23& Training was cumbersome for this system. [-] & -0.38 & -0.15 \\
7& Did you experience delays between your actions and expected outcomes? [-] & -0.36 & -0.22 \\
11& How much did the control devices interfere with the performance of test course tasks? [-] & -0.28 & -0.20 \\
20& I imagine most people would learn to use this system very quickly. & 0.27 & 0.17 \\
18& I think that I would need the support of a technical person to be able to use this system. [-] & -0.21 & -0.35 \\
16& I found the system unnecessarily complex. [-] & -0.15 & -0.11 \\
13& How well could you localize sounds? & 0.14 & 0.26 \\
\bottomrule
\end{tabular}
    \caption{Operator Judge Survey questions and their Pearson correlation coefficients with Operator Score and Task Score.  [-] indicates questions with reverse polarity, i.e., 1 best, 5 worst. }
    \label{tab:SurveyCorrelations}
\end{table*}

\subsection{Impact of Technologies on Performance}

\subsubsection{Qualitative Observations}

We offer observations on how five main aspects of the Avatar system --- navigation, manipulation, grasping, vision, \revision{and networking} -- qualitatively affected Finals scoring. 

{\bf Navigation}
Tasks 5 and 8 assessed the importance of mobility. Of the 17 teams, only two attempted bipedal walking, which did not have successful results. In the case of iCub, it demonstrated stable walking up to Task 4, but encountered catastrophic failure after an unexpected collision. Overall, wheeled bases were better suited for the competition due to long distances traveled over flat terrain and the fact that completion time was also a ranking criterion.  Task 8 demonstrates the advantages of an omnidirectional base, as the only team that failed this task used a differential drive base. Out of the 15 teams which used wheeled bases, 11 teams used an omnidirectional one due to its ease of operation for obstacle-avoidance navigation.

{\bf Manipulation}
Only two teams used one robot arm, whereas the remaining teams used two. Tasks 4, 7, 9, and 10 were related to manipulation. Although it is possible to accomplish all four tasks using a single arm, an avatar with two arms can provide additional advantages in terms of stability,  efficiency, and familiarity to the operator. For instance, Task 9, which involves holding a heavy drill and loosening a bolt, can be accomplished more securely by using two hands, especially when activating the drill's trigger button and subsequently loosening the bolt. However, the competition did not afford a nuanced chance to examine this design choice, as the two one-armed teams were only able to achieve Tasks 4 and 5, respectively. 

Load force feedback was particularly important for Task 6, and could also be beneficial for Tasks 7 and 9. 6 of the 17 teams used a force feedback system in the form of a robotic arm. With the exception of Pollen Robotics, which came in second, and AVATRINA, which came in fourth, teams that utilized interfaces with force feedback in the form of a robotic arm had higher task performance scores. Pollen Robotics augmented their vibrotactile feedback from a VR controller with an elbow exoskeleton to provide force feedback.  AVATRINA used augmented reality to render load forces (a feature shared by NimbRo) as well as vibrotactile cues.

{\bf Grasping}
Teams with parallel jaw grippers could likely complete up to 8 Tasks. Task 9, which involves operating a drill, is almost impossible to perform with a parallel jaw gripper. Teams that used multi-finger and anthropomorphic grippers were better equipped to address Task 9 because they were able to press the drill's trigger button while simultaneously holding it. Pollen Robotics and Team Northeastern succeeded up to Task 10 with three-finger grippers, properly sized to achieve the competition tasks.

For manipulation control, most of the top teams used haptic gloves in their operator system, with the exception of the second place winner Pollen Robotics, which used a VR controller with vibrotactile feedback.

{\bf Vision} 
Vision feedback was deemed to be crucial for fine manipulation tasks, as well as situational awareness during navigation on the relatively dark course (Tasks 5 and 8). Moreover, the background behind objects in manipulation Task 6 and 9 was black, which often led to low acuity in observing the black-colored trigger of the drill and black-colored portions of robots' grippers. 

Most teams used a stereo camera system and HMD for an immersive experience and improved depth perception. Among the top-ranked teams, only Team Northeastern, placed 3rd, used a mono camera system and a screen. However, they used a large monitor to improve immersiveness and utilized an innovative laser projector to improve the operator's depth perception for manipulation and navigation tasks. 


\revision{{\bf Networking}
The WiFi network at Finals did not perform consistently within the large testing arena, so the maturity of Teams' networking and communication infrastructure had a major impact on reliability and speed. During ``warm up'' testing, network connectivity issues at the start of the course forced testing to be delayed many hours past the planned schedule.  After organizers reconfigured the network and teams reconfigured internal avatar settings, teams were able to begin their scored runs, although some teams were affected by intermittent communication failures during their runs that incurred delays, especially if their system required a manual remote restart after disconnection. }

\subsubsection{Quantitative Analysis}

Here we examine correlations between select technology usage -- both on the Avatar Robot and Operator System -- vs outcomes, namely Task score, Judge score, and judge survey ratings.  We used technology usage indicator variables for omnidirectional base; differential drive base; legged base; bimanual manipulation; maximum finger count per hand; head-mounted display (HMD) usage in HMI; glove usage in HMI; and arm force feedback usage in HMI. Because only one team used legs exclusively, we include legged robots with wheels in the definition of a legged base.  Maximum scores / ratings for the runs are used as the target value for each outcome variable. \revision{We chose a $p<0.1$ significance level, which is larger than the ``usual'' 0.05 threshold due to the small population size.}  Using a linear fixed effects model~\cite{FixedEffectsFARKAS200545}, isolating each technology usage indicator against each outcome variable, we report the following interactions significant:
\begin{itemize}
    \item Omnidirectional base is positively associated with Q4 (anticipating results, effect estimate 1.07) and Q14 (manipulation proficiency, EE 1.30).
    \item Differential drive base is negatively associated with Q4 (anticipating results, EE -0.87).
    \item Legged base is negatively associated with Q14 (manipulation proficiency, EE -1.62).
    \item Bimanual manipulation is positively associated with Task score (EE 3.14).
    \item Glove usage is associated with Q12 (concentration on tasks, EE 0.91) and Q23 (training cumbersome, EE -1.02).
    \item Arm force feedback is positively associated with Task score (EE 3.00), Judges score (EE 0.88), Q3 (experience consistency, EE 1.38), Q8 (adjustment speed, EE 0.75), Q12 (concentration on tasks, EE 1.00), Q14 (manipulation proficiency, EE 1.13), Q19 (integration quality, EE 0.88), Q22 (confidence, EE 0.88). It is also associated with Q5, Q10, Q13 which involve vision and audio quality.
    \item Finger count and HMD are not significantly correlated with any outcome variable.
\end{itemize}
The lack of significance in many associations can be attributed to a small sample size ($n=16$). Hence, we report the following additional weakly significant interactions at the $p<0.2$ level (effect estimates in parentheses):
\begin{itemize}
    \item Omnidirectional base is weakly associated with Q6 (0.87), and Q15 (1.20).
    \item Legged base is weakly associated with Judges score (-0.64), Q1 (-0.97), Q6 (-1.08), and Q9 (-0.92).
    \item Finger count is weakly associated with Q1 (-0.33)
    \item HMD is weakly associated with Q1 (-1.08), Q7 (-1.18), Q19 (1.00), Q23 (-1.03).
    \item Arm force feedback is weakly associated with Q4 (0.75), Q9 (0.75), and Q23 (-0.75).
\end{itemize}
Although it appears that arm force feedback has the strongest association with positive scores, its correlations with seemingly irrelevant questions (Q5, Q10, Q13) suggests that use of this technology is a proxy for team sophistication and preparedness. Conversely, the use of a legged base is likely a proxy for brittleness in the system, which was reviewed poorly overall.

Interestingly, although no tasks explicitly required bimanual manipulation, teams with two arms tended to achieve much higher task scores than those with one. Part of this trend may be explained by overall team preparedness and resources, since building a robot with multiple arms is generally more time consuming and expensive than restricting the robot to one arm. Having two arms, however, also increases the robustness of the overall system, which allowed some operators to complete tasks even if one arm was disabled (for example, during Pollen Robotics' and Nimbro's runs). A bimanual robot also allows for asymmetric designs, allowing each arm's capabilities to be tuned to different tasks (this approach was used by Cyberselves, AVATRINA, and Dragon Tree Labs).

Other subtle observations include omnidirectional movement helping anticipate results compared to differential drive bases, which could be explained by differential drives requiring much greater precision in orientation control to avoid lateral misalignment.  Also, glove usage can be understood as requiring less training and cognitive overhead in controlling gripper movements as compared to joysticks and VR controllers.  This could be explained because such controllers require operators to develop accurate mental models of button mappings, whereas gloves exploit the ingrained experience of moving one's hands.

Surprisingly, teams that used screens rather than VR headsets did not seem to incur any statistically significant penalty despite the absence of stereo depth perception. This observation can be explained by other factors, e.g., screens may be considered more comfortable, can provide higher resolution imagery, and can be augmented with other depth cues, such as the approach used by Team Northeastern.




\section{Discussion and Lesson Learned}
\label{sec:Discussion}

\subsection{Judges' Observations}

Operator Judges gained experience at Semifinals and Finals on many different systems, each with their own quirks and varying capabilities. Judges therefore needed to become personally proficient in up to five systems over the two testing events.  Humans have remarkable capabilities of adaptation, and Judges largely adapted to limitations of the control systems and Avatar robots to help teams compete.  However, the ultimate success of a team was determined not just by the capability of their Avatar robot but also how well operators were trained to leverage those capabilities within the limited training time.  \revision{In this section, we summarize qualitative observations and commentary from Operator judges involved in the competition.}

\subsubsection{Mental Models and Cognitive Load}

During training, the Operator begins to construct a mental model of the Avatar's key functions, and during testing the Operator uses this model to operate the robot while augmenting it with practical experience. Several factors influence the success of an Avatar system, including intuitiveness and predictability of the sensorimotor mapping, training protocol, cognitive load, and even operator attitude.

Systems which have been designed to minimize the cognitive load of the operator, and give more time  to practice the fullness of their robotic capabilities will instill confidence in the operator. They can say, ``I understand this is what I need to know about this particular manipulator, all of the fingers don't work. But in order to do the grasping, I can manage as long as I can position the arm in a particular manner.'' If an operator understands the difference between how to perform an action in reality and how to perform it via the Avatar, this can help operators to transition from fighting the Avatar to truly feeling present in the remote location. 

Managing cognitive load through intuitive control mechanisms is essential to embodiment. Control system adjustments and limitations lead to operator fatigue and distraction, as well as mental operations required to make educated guesses where information was unclear. Some systems were so cumbersome that they caused significant exhaustion on the part of the operator. Moreover, sometimes the interface was not readily customizable to the dimensions or preferences of the operator. All of these elements create Breaks In Presence which detract from vicarious sensation.

Predictability is another important aspect of cognitive load management. Operators need to consistently understand the difference between how humans perform an action, and how it is to be undertaken using an Avatar. Reducing the variance between operator interaction input and the machine’s resulting output expression can enhance the operator’s mental model of the mapping of their actions to the avatar’s actions, and also reduce frustration and fatigue.

Effective and streamlined training for the operator judge was a key success factor, and quality of the training program was a key differentiator that helped distinguish the very top teams.  Training protocol made a huge difference in learnability and skill development with the avatars. The time provided for training was limited by the practicalities of the competition, as well as operator fatigue. Each system was different, and a Judge had to learn each system’s capabilities and limitations, along with various nuances of the system in a one hour session.  Teams that had well-prepared training protocols that exposed operators to conditions, tasks, and objects similar to those on the course tended to have better outcomes than those that did not.  Moreover, a team’s morale and attitude, and their effectiveness in communicating and collaborating, within their group, and with Operator Judges had a likely impact on outcomes. 

Cognitive load management within teams themselves was also critically important when facing the often-chaotic nature of preparing an avatar system. Troubleshooting connectivity, power, or mechanical issues that inevitably arose was managed well by some teams, and not as well by others. Some of the better-organized teams had well-developed startup protocols that integrated checklists to reduce cognitive load and ensure that desirable features have been activated. Such factors were not explicitly tracked during the competition, but remained important factors beyond the scoring that should be considered in future competitions.

The differences between robotic avatars, interaction mechanisms, and judges, all induce variability.
Judges were afforded a break in between their usage of different systems to rest, as well as to adjust and acclimate, but still faced significant challenges adapting between respective systems.

\subsubsection{Observations on Avatar Systems}

Teams took multiple approaches with their Avatar systems. Some teams competed with existing, commercially available robots, whereas others constructed systems ad-hoc to meet the requirements of the competition. The most complex robots were distinctly humanoid, at human scale, with some even attempting to replicate human ambulation. Moreover, humanoids have relatively high restrictions in terms of weight and integration. Generally, the more complex and humanoid avatar systems were at significantly elevated risk of catastrophic and irrecoverable failure. 

Conversely, several systems which seemed less impressive in their stature or capabilities often performed reasonably well despite their limitations. This resiliency may have been due to relative ease of control for the operators, who were therefore able to work around any physical constraints. This provides a lesson for non-autonomous robotics, that there is an optimal level of complexity. The avatar robots that performed the best were those that were customized and adapted to meet the needs of the competition precisely, without exceeding them. Systems which attempted to provide a solution for every imaginable problem or situation may ultimately suffer due to over-complication.

Ensuring system robustness in a wide range of conditions was important for team success. As an example, one team (Tangible/Converge) experienced a spring-loaded connection failure. When an Operator happened to flex their arms in a manner that felt natural for them during communication with the Mission Commander, an unanticipated snag occurred, unplugging a connector with the robot arm. The robot was unable to recover the arm or reboot the system, which prematurely ended the team's run.

\subsubsection{User Interfaces and Situational Awareness}

User interaction and interface elements are a crucial aspect of functional robotic telepresence, as they are key to enabling human proficiency. Judges reported that this seemed to be an underdeveloped aspect of many teams, who focused more on core engineering. In contrast, teams that succeeded in the competition focused heavily on user experience. This highlights a recommendation for robotic avatar engineering teams to include experts in human-machine interaction, user experience, and human factors to ensure that the user experience is not overlooked. 

User interfaces also presented challenges. Most teams adapted commercial Virtual Reality systems to their respective robots. These could include visual indicators of the parameters of the robot and its environment, such as the weight of a payload in the arms, or navigational aids. However, the display of such information required finesse in order to not overwhelm the operator. Some teams chose to implement modal settings, which changed the information provided, as well as the finesse of control, according to the present task. However, some of these implementations were rather rough, basic, and presented persistent error codes, or required an operating system restart to switch tasks. 

Operators could request feedback or assistance from the avatar team, but maintaining conversation whilst in the midst of piloting tasks was very challenging. Another important element was providing an ability to reset the system to the operator, so that they could hopefully fix an issue quickly rather than relying on outside assistance.

Many of the errors in manipulation were because of the spatial doubt caused by a degraded visual experience. The narrow field of view of video feeds provided by cameras and VR systems presented challenges. In some cases, this led to collisions due to a lack of perception of the environment around the avatar robot, and also provided insufficient visual feedback to aid proprioception of the arm manipulators. Many systems also had video feeds with a poor refresh rate, or worse, distinguishable lag time. Lag times of multiple seconds were often the root cause of failure of the course which was insufficient for the operator to see effectively through the unit. 

Other safety features, such as a rear-view camera upon backing up a robot, also played an important role in the avoidance of accidents. One team experienced a catastrophic crash on the course upon backing over a rock. With the aid of cameras this might otherwise have been avoided, especially with the addition of collision sensors which could automatically interrupt hazardous operator inputs.

\subsubsection{Lessons Learned from Top Teams}

The top-scoring system NimbRo~\cite{SchwarzIROS21, SchwarzNimbRo2022} from University of Bonn,  Germany had particularly advanced technology, especially on the operator side. This team built a full replica of the avatar’s robotic arms for the operator to manipulate. Operators therefore reported that there was no ``mapping'', as whatever action they took on the control arm was mapped directly to the robot, resulting in a highly  vicarious experience. They also had a six degrees of freedom articulated camera/face screen combination, enabling the operator to view the scene from different angles for a spatial sense of the environment, in a natural analog to human head and neck movements. This team’s user interface was also well honed. When the robot backed up, a rear camera was displayed, and then hidden again when not in use. 

The second place winner, Pollen Robotics from Bordeaux, France developed an open source system which used commonly available off-the-shelf hardware [11]. The remote avatar featured significant 3D printing, reducing costs in an elegant manner. Consumer grade VR headsets and controllers provided a simple interface instantly familiar to computer gamers and allowed rapid adaptation to the context. When the robot was driving, the LiDAR map grew to be more visible, and then automatically shrunk whilst in manipulation mode. Operators reported that the robot just ``did what you wanted them to do", hiding the complexity of the robot behind an intuitive interface that obscured complexity. Despite this, the system did not feature much in the way of semi-automated processes, such as navigation and obstacle avoidance, object detection, or grasping routines. Whilst this necessitated more manual control by the operator, it sidestepped the potential for error on the part of the system. Other systems featured significant task oriented modes which facilitated performing certain actions.

\begin{figure*}[tbp]
    \centering
    \includegraphics[width=0.96\linewidth]{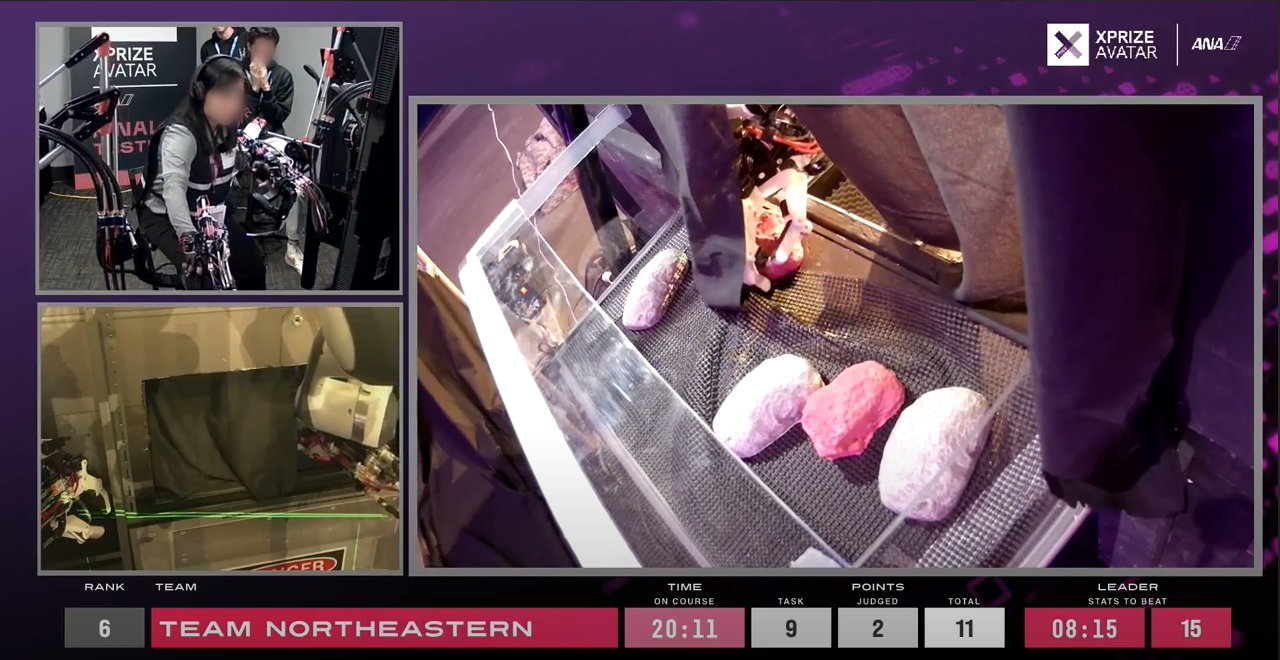}
    \caption{The rock identification task during Team Northeastern's successful run. Scratching sounds replayed over audio combined with rudimentary haptic feedback while the rock was touched with the robotic hand led to successful identification of the rough rock.  }
    \label{fig:FinalNortheasternRocks}
\end{figure*}

Top-scoring teams included more thoughtful user-interface design choices that not only helped reduce cognitive load. One example of such a design choice was adaptive user interface components, which either appeared and disappeared based on operator needs, or changed their appearance dynamically. NimbRo’s user interface automatically switches camera views based on the movement direction of the robot, giving the operator the most useful visual feedback for each situation. Pollen’s user interface featured a dynamically adjusted map scale, showing a broader map when the robot was moving fast, and a more zoomed-in map when the robot was moving slowly, minimizing the map when the robot was in manipulation mode. These relatively simple, but thoughtful, features reduced operator cognitive load and helped them focus on the task at hand. 

Top-scoring teams also built in methods to help the operator judge compensate for the robot’s limitations. 
This insight was applied in the design of the operator interface of the Team Northeastern robot.  Figure~\ref{fig:FinalNortheasternRocks} captures a moment during the last task of the competition.  The objective of this task is to use the robot hand to feel the rocks and select the one with a rough texture.  As can be seen in the upper left screen of Figure~\ref{fig:FinalNortheasternRocks}, the hydraulic system provides very minimal resolution for fine motor feedback. Scratching the surface of the rock cannot, for example, be felt on the operator’s fingertips.  Further, the lower left screen shows the visual view of the operator.  The black curtain completely occludes her view, thus she cannot see the textures of the rock. Note that the right view is displayed to audience members only, and is not available to the operator.  Yet, the operator successfully identified the rough rock.  The Northeastern team installed a microphone near the hand, which produced a scratching sound when the robot hand moved over the rock surface.  Despite a lack of advanced high resolution haptic feedback and the complete lack of visuals, the additional audio cue, combined with the proprioception for the movement of the operator’s own hand provided the necessary compensation to make the correct choice. While there is a need to continue advancing each modality, designers may also consider a holistic, multi-modal approach to improve operator experience and effectiveness.

\subsection{Limitations of Competition}

Several limitations and criticisms of the ANA Avatar XPRIZE competition have been raised by teams and outside observers.  Broadly, these fall into the categories of 1) the change of scope from semifinals to finals creating a moving target and de-emphasizing social interaction, 2) the evaluated tasks being of low complexity and encouraging overfitting, and 3) low gender diversity and bias toward high-resource teams.

Between semifinals to finals, the competition was transformed into a show, and this push biased the entire scenario definition of the finals. Change of scope and context (room vs Mars) required higher complexity in defining objective metrics for embodiment and one may wonder if human-like avatars were necessary in the Mars context.  The competition schedule was very tough, with only 5 months between the release of the finals rules and the competition itself. International teams had to spend several weeks shipping their systems.  

The original purpose of the competition was to advance  telepresence, the sensation (for one's self) and expression (for observers) of being in another place. Carrying out practical operations in an efficient manner is part of telepresence, but the Finals competition arguably failed to meet the original purpose of the competition because it overemphasized manipulation and navigation tasks with very little social interaction. This failed to encourage teams to consider technologies like full-body touch sensing and haptics which would be critical for many social applications of avatars.  By limiting the scope of the Finals, the competition promoted simplicity of avatar systems rather than full embodiment capabilities.


The nature of a task-based competition naturally encourages task-specific solutions. This was minimized to some degree by withholding the specific details of the Final's competition tasks to months before the competition and instead releasing general guidelines and information on the types of tasks to be performed (grasping, haptic perception, etc.). Nonetheless, several teams did manage to develop application-specific hardware and software for the more challenging tasks such as drill operation or texture identification. In future competitions, to encourage general-purpose technology it may be advisable to withhold task details until just a few days before the competition, or to include an overabundance of possible tasks to complete so that teams would be forced to develop more general solutions. In such a setting, a competition should not be structured around a linear high-stakes trial as in Finals, because it would penalize teams that neglected to perfect particular technology area early in the trial. Rather, point accumulation across tasks (like a decathlon or gymnastics competition) would be a preferable structure.

It can also be observed that the tasks tested in the Finals competition are significantly simpler than those that would be encountered in relevant telepresence application domains, which may require fine dexterous manipulation, mobility over rough terrain, full-body touch sensation, augmented perception with superhuman sensory modalities, or direct and intimate physical human interaction~\cite{IEEESpectrum2018b}. This criticism was levied at XPRIZE's proposed tasks by external observers early in the stage of the competition in 2018~\cite{IEEESpectrum2018}, and yet the tasks tested in 2022 were ultimately even simpler than those proposed earlier.  Avatars were required to manipulate a small number of objects, which were pre-specified. The locomotion task was on flat terrain without significant obstacles, so almost all solutions converged towards similar wheeled robot form. The competition also eliminated many sensor modalities, e.g., temperature (both ambient and object), humidity, airflow, smell), which were mentioned in early stages of the XPRIZE competition but eliminated in Semifinals and Finals. 

Another criticism of the competition was barriers to participation. While the Avatar XPRIZE Finals had globally diverse participants from 16 countries across 4 continents, the most successful teams were affiliated with well-established and well-funded research labs in academia and industry. The complexity of designing an avatar system requires numerous skills as well as access to technology and materials.  Several lesser-funded teams did not advance to the physical testing phase of the competition. There were also apparent barriers to achieving gender diversity on the teams. Of the seventeen finalists, eight appeared to have all-male teams, and five of the remaining nine teams included only a single woman on the team.  This continues to be a challenge for many STEM competitions.

\subsection{Impact of Competition on Telerobotics and Telepresence}

In terms of impact to the broader field of telepresence, the ANA Avatar XPRIZE competition provided momentum in a number of ways. As noted in IEEE Spectrum~\cite{IEEESpectrum2023}, ``the competition provided the inspiration (as well as the structure and funding) to help some of the world’s best roboticists push the limits of what’s possible through telepresence.'' Over the course of the competition, hundreds of team members across the globe were focused on developing novel Avatar technologies while also raising over 50M dollars in team investments. The competition elevated the conversation around avatars among business, tech and media while simultaneously fostering collaboration across multiple disciplines to integrate technologies.

Participants contributed to academic journals and conferences, including this special issue.  Several workshops were also organized with ties to the competition:
\begin{itemize}
    \item IEEE Future Directions Workshop 2021 ``The Future of Telepresence''\footnote{\url{https://ieeetv.ieee.org/event/ieee-telepresence-workshop}}
    \item Robotics: Science and Systems (RSS) 2022 workshop ``Toward Robot Avatars: Perspectives on the ANA Avatar XPRIZE Competition''\footnote{\url{ https://www.rssavatarxprizews.org/program.html}}
    \item European Robotics Forum 2022 workshop ``A Future with Avatar Robots''.
    \item IEEE Symposium on Telepresence 2022\footnote{\url{https://telepresence.ieee.org/events/2022-ieee-symposium-on-telepresence}}
    \item The upcoming ``2nd Workshop Toward Robot Avatars'' at IEEE International Conference on Robotics and Automation (ICRA) 2023
    \item Future of Telepresence Workshop at IEEE Systems, Man, and Cybernetics Conference 2023
\end{itemize}

Long-lasting initiatives related to the XPRIZE competition include the XPRIZE alumni network and IEEE Future Directions project on Telepresence\footnote{\url{https://telepresence.ieee.org/}}. IEEE Telepresence was launched in 2021 as a community for projects, events and activities on telepresence technologies, including drafting a roadmap for future research.

Furthermore, the competition and its associated research raises awareness of the need for evaluation infrastructure development in telepresence systems. In order to quantify telepresence performance, evaluators must measure subjective criteria—like sensations of immersion, embodiment, presence, and cognitive load. The standard methodology used throughout human-robot interaction research is to conduct user studies followed by subjective questionnaires, but it is extremely hard to apply this methodology consistently to compare performance across Avatar systems. Testing may be geographically dispersed, tested by different operator populations, with differing environmental and networking conditions, and with different levels of operator training and preparation.  The XPRIZE competition standardized at least some of these factors to provide a comparative snapshot of avatar systems at this point in time. However, to further nurture progress in telepresence research it would be immensely helpful to standardize evaluation procedures, including but not limited to benchmark tasks, survey questions, training duration, and environmental conditions.

Telepresence research increasingly includes modalities beyond vision, audition and haptics such as temperature~\cite{drif2005iros,guiatni2011jdsmc,Fermoselle2022}  and social touch~\cite{vanHattum2022}.  One of the difficulties that limits the ``multimodality'' of telepresence systems is the absence of reliable mobile sensor systems for some sensory modalities such as smell and (whole body) delicate touch. Also, most systems are restricted to render sensory cues directly from sensor data, limiting their functionality to the sensors that are present in the remote environment. Semantic analysis of sensor data is a possible fix, for instance the generation of a coffee smell based on recognition of the size and form of a cup, the color of the fluid it contains, and the words spoken by someone holding it. Semantic-based cue generation is a developing field and only a few references started to appear in literature in recent years~\cite{GarciaPereira2020,Daiber2021}.

It can be observed that in comparison to NimbRo's winning time of $\sim$6 minutes, a human performing these tasks in-person could have likely finished the course within 1--2 minutes, depending on the perceived urgency.  While the result is technically impressive, it signifies that there is still a significant gap between the current state of Avatar technology and the ultimate goal of matching (or exceeding) human performance via telepresence.  There are two general directions for overcoming this gap: continue to improve immersiveness and transparency so an operator feels closer to using their own body, or add shared control capabilities to aid the operator. The former approach faces fundamental problems of time delay and robot-human mismatch, while the second approach reduces immersion and predictability of the control mapping, and hence typically requires more training.

\section{Conclusion}
The ANA Avatar XPRIZE competition was a milestone in the field of telepresence. Unified by a common goal of enabling novice operators to sense, communicate, and act in a remote environment, the competition inspired many engineers, judges, and observers to make sustained advances to the field.  A variety of technical approaches were employed, combining robotics, virtual and augmented reality, and haptics, to convey sensation, navigation, communication, and dexterity between the operator and remote robot. Immersiveness, transparency, and fidelity were important factors to keep training time for operators to a minimum.  We analyze the results of the competition to identify the most important factors for telepresence success, including a strong association between objective task success and subjective operator impressions, the need for human-like arms and grippers and high-resolution vision, and limiting cognitive load through intuitive operator interfaces.

The competition also raises many challenges for the future of telepresence. On a technical level, more complex tasks will demand high-fidelity haptics for fine manipulation, control of dynamic movement, navigation over non-flat terrain, full body haptics, and use of other sensor modalities such as temperature and smell.  Organizationally, more work is needed to sustain avatar research, such as standardizing evaluation techniques and infrastructure for systems that are geographically and temporally distributed, and to develop standardized platforms that can scale to large numbers of users and robots.  

\section*{Declarations}

Authors include representatives of ANA Avatar XPRIZE teams (K. Hauser, P. Naughton, AVATRINA; J. Bae, Team UNIST; S. Behnke, NimbRo; M. Catalano, AlterEgo; S. Dafarra, D. Pucci, iCub; J. van Erp, i-Botics; J. Fishel, Converge Robotics; F. Kanehiro, A. Kheddar, Janus; G. Lannuzel, S. Nguyen, Pollen Robotics; P. Oh, J. Vaz, Avatar Hubo; T. Padir, P. Whitney, Team Northeastern; J. Park, Team SNU), judges (E. Watson, J. Bankston, B. Borgia, T. Ferris, G. Hoffman, S. Ivaldi, J. Morie, P. Wu), and organizers (J. Pippine, D. Locke).

\paragraph{Financial Interests} No funding was received to prepare this manuscript. 

\paragraph{Conflicts of Interest}
Some authors are representatives of ANA Avatar XPRIZE teams that received award money from the XPRIZE Foundation as a result of their teams' performance during Semifinals and Finals (K. Hauser, J. Bae, S. Behnke, J. Fishel, G. Lannuzel, P. Naughton, S. NGuyen, P. Oh, T. Padir, J. Park, and P. Whitney). Authors J. Pippine and D. Locke were employed by XPRIZE Foundation during the ANA Avatar XPRIZE competition.

\paragraph{Consent to Publish} XPRIZE judges signed informed consent regarding publishing their data and photographs.

\section*{Data Availability}

The data that support the findings of this study are available from XPRIZE Foundation but restrictions apply to the availability of these data, which were used under licence for the current study, and so are not publicly available. Data are however available from the authors upon reasonable request and with permission of XPRIZE Foundation.

\clearpage
\bibliographystyle{plain}
\bibliography{ref,ref-background,ref-technologies}

\end{document}